\documentclass[runningheads]{llncs}

\usepackage{multirow}
\usepackage{appendix}

\usepackage[mobile]{eccv}

\usepackage{eccvabbrv}

\usepackage{graphicx}
\usepackage{booktabs}

\usepackage[accsupp]{axessibility}  

\usepackage[pagebackref,breaklinks,colorlinks,citecolor=eccvblue]{hyperref}
\usepackage[utf8]{inputenc}
\usepackage{orcidlink}

\begin{document}

\title{Rethinking Referring Object Removal} 

\titlerunning{Rethinking Referring Object Removal}

\author{Xiangtian Xue\inst{1,3}\and
	Jiasong Wu\inst{1,3}\thanks{Corresponding author}\and
	Youyong Kong\inst{1,3} \and
	Lotfi Senhadji\inst{2,3} \and
	Huazhong Shu\inst{1,3} }

\authorrunning{X. Xue et al.}

\institute{Key Laboratory of New Generation Artificial Intelligence Technology and Its Interdisciplinary Applications, Southeast University \and
Laboratoire Traitement du Signal et de l'Image, Univ Rennes \and
Centre de Recherche en Information Biom\`{e}dicale Sino-fran\c{c}ais (CRIBs)\\
\email{xxt@seu.edu.cn}, \email{jswu@seu.edu.cn}}

\maketitle

\begin{abstract}
  Referring object removal refers to removing the specific object in an image referred by natural language expressions and filling the missing region with reasonable semantics. To address this task, we construct the ComCOCO, a synthetic dataset consisting of 136,495 referring expressions for 34,615 objects in 23,951 image pairs. Each pair contains an image with referring expressions and the ground truth after elimination. We further propose an end-to-end syntax-aware hybrid mapping network with an encoding-decoding structure. Linguistic features are hierarchically extracted at the syntactic level and fused in the downsampling process of visual features with multi-head attention. The feature-aligned pyramid network is leveraged to generate segmentation masks and replace internal pixels with region affinity learned from external semantics in high-level feature maps. Extensive experiments demonstrate that our model outperforms diffusion models and two-stage methods which process the segmentation and inpainting task separately by a significant margin.
\end{abstract}

\section{Introduction}

Traditional image inpainting techniques~\cite{Wang_2022_CVPR} perform well in filling missing regions based on an input mask, but there often exist vague distinctions between inpainting background features (\ie object removal) and foreground features. For example, when a person's eyes are masked, it is difficult for models to discern whether the user intends to remove the eyes entirely or generate new ones. The introduction of text-based instruction will effectively address this issue by allowing users to explicitly declare their desired manipulation.

This paper further specifies text-instructed inpainting as the Referring Object Removal (ROR) task, which aims at removing the specific object in an image according to natural language expressions. Language-based instruction serves as the most fundamental and convenient interactive modality in real-world scenarios, such as chatting with a robot commonly using language as the sole mode of interaction. Another typical application scenario is performing similar manipulations on extensive datasets such as image desensitization and video pixelation. With a simple descriptive sentence, users can simultaneously eliminate objects with the same attributes from numerous visual entities.

To the best of our knowledge, scholars have done little research in this area (see \cref{2.1}), and there is a lack of professionally labeled datasets. To bridge this gap, we construct ComCOCO, the first semantic removal dataset that consists of images with referring expressions and corresponding ground-truth results after elimination. The synthetic dataset facilitates the model to attach more importance to the structure of objects than image details~\cite{2018Training} and provides diversiform coordinated visual combinations based on semantic scenes in RefCOCO+~\cite{2016Modeling}. The construction process endeavors to rationalize the context semantics of composite images and concurrently augments random noise which heightens the robustness of the model.

Furthermore, we propose an end-to-end baseline model called Syntax-Aware Hybrid Mapping Network (SAHM). With an encoding-decoding structure of the input image, hierarchical text features extracted at the syntactic level are fused with the whole downsampling process of visual features. The upsampling process produces hybrids of segmentation mapping and inpainting mapping. The segmentation pathway combines multi-scale semantic features, followed by spatial alignment with deformable convolution to regulate the mask boundary. With the location information of segmentation feature maps, the inpainting mapping further learns region affinity between patches inside and outside mask regions and fills in with coherent semantics. The main contributions of our work are highlighted as follows:
\begin{itemize}
	\item We comprehensively explore a novel multimodal task named Referring Object Removal that aims at removing specific objects in a given image guided by natural expressions and filling in with reasonable visual semantics.
	\item We construct the synthetic dataset ComCOCO containing 136,495 referring expressions for 34,615 objects in 23,951 image pairs that consist of images with descriptive expressions and the corresponding ground-truth images after elimination. 
	\item We further propose an end-to-end network with hierarchical text-guided visual attention at the syntactic level and hybrid inpainting with the location information of the segmentation mask. Empirically, we demonstrate that our end-to-end method outperforms existing approaches by a significant margin.
\end{itemize}

\section{Related Work}
\subsection{Diffusion-based Image Editing}\label{2.1}
Diffusion models have demonstrated impressive generative capabilities in Text2img tasks, and recent researches~\cite{brooks2023instructpix2pix,huang2023smartedit} have extended this capacity to image editing guided by textual instructions, where the editing scenes predominantly revolve globally altering the style of an image based on simple scenes. The most relevant study is Inst-Inpaint~\cite{yildirim2023instinpaint} which focuses on instructing to remove objects. However, the ground-truth images within its proposed dataset are segmented and inpainted using existing methods, which significantly diminishes its validity and potential contribution to the community. The removal performance exhibited in its paper is also unsatisfactory. Therefore, this paper serves as a supplement to the research area with a new perspective, and further analysis of diffusion models is elucidated in \cref{5.3}.
\subsection{Referring Image Segmentation}\label{2.2}
Referring image segmentation aims at segmenting the visual entity referred by natural language expressions. Existing methods can be roughly divided into two categories: one-stage methods that directly mask the object via fusing semantic and visual information~\cite{Wang_2022_CVPR,yang2023semanticsaware}, and two-stage methods that first propose instances and then select the optimal candidate supervised by linguistic features~\cite{Luo_2020_CVPR}. Transformer~\cite{NIPS2017_3f5ee243} has been proven effective for this task these years. Ding et al.~\cite{Ding_2021_ICCV} introduce Transformer and multi-head attention to construct an encoder-decoder attention mechanism architecture. LAVT~\cite{Yang_2022_CVPR} achieves better cross-modal alignments through the early fusion of linguistic and visual features in a vision Transformer encoder network. Nevertheless, the authors merely combine Swin-Transformer~\cite{Liu_2021_ICCV} with EFN~\cite{Feng_2021_CVPR}, and the performance improvement intrinsically reckons on the selection of the backbone. 
\subsection{Image Inpainting}\label{2.3}
Image inpainting refers to the process of leveraging the surrounding auxiliary information and filling in missing pixels of an image. Recent learning-based methods introduce contextual attention to transfer background features to missing regions~\cite{song2018contextual,wang2021image}. Zeng et al.~\cite{zeng2019learning} propose a pyramid-context encoder network that progressively learns region affinity by attention from a high-level semantic feature map. Guo et al.~\cite{Guo_2021_ICCV} develop contextual feature aggregation module to refine the generated contents by region affinity learning and multi-scale feature aggregation. Cao et al.~\cite{cao2022learning} utilize group convolutions to aggregate CNN features with attention scores. Most inpainting tasks aim at restoring foreground features to ensure consistency with the original image, while we emphasize the elimination of foreground features and restructure of the obscured background.
\subsection{Datasets}\label{2.4}
Current object annotations of referring datasets mainly rely on two-player games and automatic generation technology~\cite{2016Modeling,wu2020phrasecut,mao2016generation}. The image source is usually taken from MSCOCO~\cite{lin2014microsoft}. Meanwhile, the datasets appropriate for image inpainting differ in diversified image scenes, and the common practice of constructing missing regions is to append irregular holes in images~\cite{7968387,liu2015deep,tylevcek2013spatial}. For the ROR task which emphasizes the elimination of complete objects with linguistic annotations, it is arduous to manually annotate existing public images restricted to the indetermination of representation under the occlusions. In ideal circumstances, researchers can snap a picture, fix the camera angle, remove the object in the lens and then snap the ground truth, but even a slight camera shake or illumination variations would be disastrous.
\section{ComCOCO}
\subsection{Dataset Creation}\label{3.1}

RefCOCO+~\cite{2016Modeling} is a referring expression dataset consisting of 141,564 expressions for 49,856 objects in 19,992 images, where expressions are strictly based on appearance without spatial information. Benefiting from RefCOCO+, we leverage its images and corresponding expressions to construct our ComCOCO dataset. Specifically, two images with similar scenes and disparate objects are selected, then the object in one image is harmoniously placed in the other image, which will not cause spatial ambiguities and conflicts owing to the unique characteristics of RefCOCO+. The raw background image is deemed as the ground truth of removal. The overall construction process is shown in~\cref{fig:onecol7} and elaborated as follows:

\begin{figure*}[tb]
	\centering
	\includegraphics[width=1\linewidth]{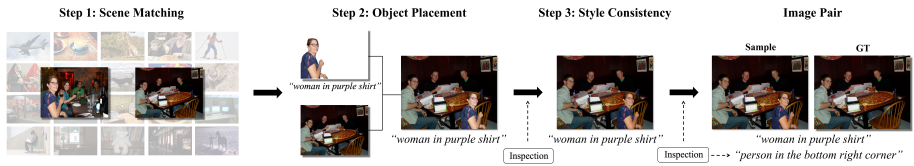}
	
	\caption{The construction process of the ComCOCO dataset, which contains Scene Matching, Object Placement and Style Consistency. The two selected images have similar semantic scenes, while the characteristics of description statements, that is, the morphological characteristics of objects, are completely disparate. Step 2 and Step 3 ensure the rationality of splicing to the greatest extent. Processed image pair contains an image with referring expressions and the ground truth. We conduct dual-phase manual inspection in the pipeline to guarantee image fidelity and incorporate spatial-based descriptions.}
	\label{fig:onecol7}
\end{figure*}

\textbf{Step1: Scene Matching.} Intuitively, any object and image can be combined to generate a noisy image, whereas it often occurs that synthetic images are incongruous and out of real life. For example, a refrigerator suddenly appears among the elephants. The reason for this phenomenon can be attributed to the semantic dissimilarity of ``refrigerator'' and ``elephant''. Therefore, we assign images in RefCOCO+ into different clusters (enumerated in the supplement) according to the category of contained objects to ensure that semantically similar images are assigned together, and vice versa. In each cluster, Jaccard similarity coefficient~\cite{2013Using} is adopted to measure the sentence similarity between the added object and other images, and five images with the lowest similarity are selected as background candidates. This step rationalizes image semantics and circumvents potential conflicts of referring expressions.

\textbf{Step2: Object Placement.} For foreground-background pairs selected in Step1, We exploit object placement technologies (\ie GracoNet~\cite{10.1007/978-3-031-19790-1_23}) to place the foreground object in the appropriate position of the background image and generate a temporary composite image. Further, we check the rationality of image composition and manually regulate the size and position of the object for unreasonable splicing. Extremely discordant and semantically conflicting scenes are directly deleted.

\textbf{Step3: Style Consistency.} There often exists an obvious disharmony in composite images due to differences in shooting angle, ambient lighting, etc. To alleviate this problem, the image harmonization method (\ie Harmonizer~\cite{Harmonizer}) is utilized to adjust the properties of foreground objects and make them compatible with the background. The pixel discrepancy between the foreground and background is eliminated after the process of harmonization, which makes composite images visually realistic and introduces challenge for the ROR task. We further recheck the integrity of referring sentences and describe spatial attributes of the specific object based on new scenario. Ultimately, composite images with referring expressions and background images as the ground truth are integrated to form the proposed synthetic dataset.
\subsection{Statistics}
The ComCOCO dataset consists of 136,495 referring expressions for 34,615 objects in 23,951 image pairs. Experimentally, we split 20,324 image pairs for training, 1,843 for validation, and 1,784 for testing. Each image pair contains an image with descriptive expressions and the ground truth after elimination. The placement and harmonization process of dataset creation greatly enhances the robustness of the model. Following MSCOCO's measurement standard for object size~\cite{lin2014microsoft}, ComCOCO contains about 11\% small objects, 53\% medium objects, and 36\% large objects. Compared with RefCOCO+ (of which the corresponding ratio is 42\%, 34\%, 24\%), our dataset increases the proportion of medium and large objects to emphasize the completion of background information, which introduces more challenges for the ROR task. More detailed statistics are provided in \cref{appendixB}.

\section{Method}
\Cref{fig:onecol1} illustrates the overall end-to-end architecture of proposed Syntax-Aware Hybrid Mapping (SAHM) network, where the input is an image with a referring expression for the specific object to be eliminated and the output is the ground-truth image. Specifically, the proposed SAHM network consists of multi-layer syntax-aware visual attention modules and hybrid mapping filling modules. Visual features of the input image are fused with hierarchical syntax-aware visual attention layers, which enhances the saliency of directional information in the language. In the hybrid mapping filling layer, the segmentation mask is further transformed into pixel holes and contextual semantics learned form region affinity between patches are filled in with hybrid mapping. 
\begin{figure*}[t]
	\centering
	\includegraphics[width=1\linewidth]{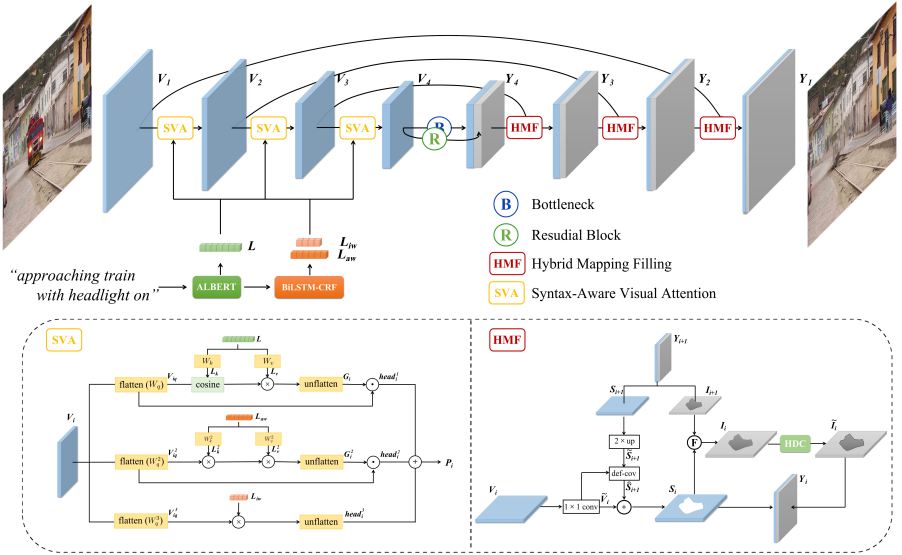}
	
	\caption{The basic learning framework of the proposed SAHM. The overall framework adopts the encoding-decoding structure based on Swin-Transformer. Hierarchical linguistic features $L$, $L_{aw}$, and $L_{iw}$ are fused in successive stages of visual downsampling via syntax-aware visual attention. $Y_i$ consists of parallel segmentation feature map $F_i$ and inpainting feature map $I_i$, which are initialized by two Swin-Transformer blocks as the bottleneck and four residual blocks, respectively. In the hybrid mapping filling module, $S_{i}$ is mapped from $S_{i+1}$ and $V_{i}$ with the skip-connection and the mask region in $S_{i}$ is further filled with external semantic mapping.}
	\label{fig:onecol1}
\end{figure*}
\subsection{Syntax-Aware Visual Attention}\label{4.1}

\begin{figure}[htbp]
	\centering
	\includegraphics[width=1.0\linewidth]{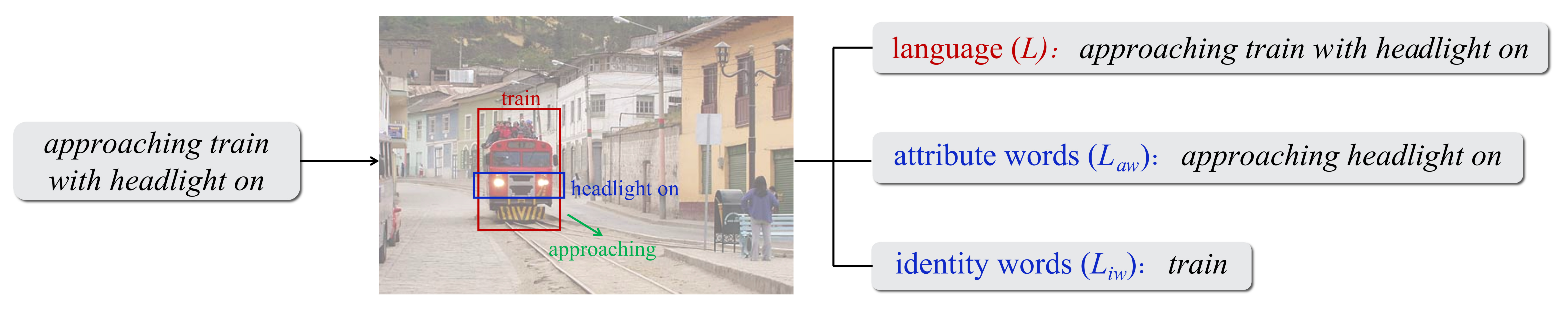}
	
	\caption{The ``\textit{identity words}'' point to the segmented object and are the core words in a sentence. The ``\textit{attribute words}'' contain attribute information including location, appearance, etc. The remainder with no representative information is deemed redundant.}
	\label{fig:onecol22}
\end{figure}

For a given image and the descriptive expression, we first extract visual and linguistic representations by modality-specified feature extractors (\ie Swin-Transformer~\cite{liu2021swin} and ALBERT~\cite{DBLP}), denoted as $V_i$ and $L$, respectively. Considering the diverse relevance to the object in language blocks, the BiLSTM-CRF~\cite{lample-etal-2016-neural} model is introduced to further capture contextual information and hierarchical dependencies at the syntactic level. Each word in the sentence is classified as ``\textit{identity words}'', ``\textit{attribute words}'', and other irrelevant words according to its directive property. For example, as shown in~\cref{fig:onecol22}, ``\textit{train}'' is the identity word since it directly describes the category information of the object. The ``\textit{approaching}'' and ``\textit{headlight on}'' are more informative than ``\textit{with}'' for the description of attribute information.

For text embedding $ L\in R^{C_L \times T_L} $, we obtain identity words embedding $ L_{iw}\in R^{C_L \times T_{L_{iw}}} $ and attribute words embedding $ L_{aw}\in R^{C_L \times T_{L_{aw}}} $, where $C_L$ refers to the dimension of vector, and $T_L$, $T_{L_{iw}}$, $T_{L_{aw}}$ refers to the length of text, identity words and attribute words, respectively. The hierarchical linguistic feature $L$, $L_{aw}$, and $L_{iw}$ integrate with visual feature map $V_i$ via syntax-aware attention modules, and the outputs are added by learnable weights to obtain the multimodal representation.

\textbf{$L$-Guided Attention.} We apply dot product on text embedding $L$ and visual feature map $ V_i\in R^{C_i \times H_i \times W_i} $ by a 1$\times$1 vector to obtain a learnable representation, where $C_i$ refers to the dimension of the $i$-th layer, and $H_i$ and $W_i$ correspond to the size of the image. The last two spatial dimensions of $V_i$ are unified into one dimension according to rows:

\begin{equation}
	L_{k}= W_kL,\  L_{v}=W_vL,\ V_{iq}=flatten(W_{q}V_i),
\end{equation}
where $ L_k, L_v\in R^{C_i \times T_L}$, $V_{iq}\in R^{C_i \times H_iW_i} $, and $W_k$, $W_v$, $W_q$ are implemented as a 1$\times$1 convolution with $C_i$ number of output channels.

We adopt cosine similarity instead of dot product to calculate the similarity between two modal features. In experiments, we find that the dot product operation renders the scope of attention dominated by certain attribute words while cosine similarity which measures in direction relatively concentrates on the whole sentence. The attention score of $V_{iq}$ for $L_k$ can be obtained as follows:
\begin{equation}
	S(V_{iq},L_{k})=\frac{cos(V_{iq},L_{k})}{ \gamma } =\frac{V_{iq}^TL_{k}}{ \gamma ||V_{iq}|| \cdot ||L_{k}||},
\end{equation}
where $\gamma$ is a learnable parameter but not shared between layers. The text feature fused with visual feature is described as follows:
\begin{equation}
	G_{i}=unflatten(softmax(S(V_{iq},L_{k}))L_v^T),
\end{equation}
where $G_{i}\in R^{C_i\times H_i\times W_i}$ has the same shape as $V_i$. Then we apply Hadamard product~\cite{Kim2017} on ${V_i}$ and $G_{i}$ to obtain the multimodal feature map $head_{i}^1$:
\begin{equation}
	head_{i}^1= V_i\odot G_{i}.
\end{equation}

\textbf{$L_{aw}$-Guided Attention.} $L_{aw}$ contains extracted attribute words in $L$ and the contribution of this layer is to sensitize the model to certain attributes to better distinguish between different objects. The structure follows the above layer and dot product is adopted to calculate the similarity of $L_{aw}$ and $V_i$. The similarity weight demonstrates a truncated distribution, where attribute words with little discrimination are suppressed and distinguishable ones are valued. 

\textbf{$L_{iw}$-Guided Attention.} The length of $L_{iw}$ is relatively short and usually contains one or two words. Thus, it is redundant to calculate the attention weight meticulously and each word is of equal importance, which is equivalent to applying Hard Attention~\cite{pmlr-v37-xuc15} on $L$ and the weight of $L_{iw}$ is set to 1.

Ultimately, the outputs of three attention heads are merged by element-wise addition with learnable weights and obtain the final output $P_i\in R^{C_i\times H_i\times W_i} $.
\begin{equation}
	P_i=Add(W_{i}^1head_{i}^1,W_{i}^2head_{i}^2,W_{i}^3head_{i}^3).
\end{equation}

\subsection{Hybrid Mapping Filling}

Upsampling feature map $Y_i$ consists of segmentation feature map $S_i$ and inpainting feature map $I_i$. The former is obtained by fusing the lower-resolution feature map and corresponding downsampling feature map, where deformable convolution~\cite{Dai_2017_ICCV} is introduced to learn transformation offsets of pixels to contextually align upsampled higher-level features. The removal region is converted into pixel holes with no semantics and filled with region affinity learned from high-level semantic features. Multi-scale information is further aggregated by hybrid dilated convolutions with different rates to refine the filled features. Successive hybrid mapping constitutes each layer of the upsampling process.

\subsubsection{Aligned Segmentation Mapping.}\label{4.2.1}

Segmentation Mapping is the decoding process of $V_i$ and generates segmentation feature map $S_i$ with the information of pixel category. The spatial resolution of lower-resolution feature map $S_{i+1}$ is upsampled by a factor of 2 and the corresponding downsampling map $V_i$ is multiplied by a 1×1 convolutional layer to reduce channel dimensions denoted as $\widetilde{S}_{i+1}$ and $\widetilde{V}_i$, respectively. Considering that there exists spatial misalignment between $\widetilde{S}_{i+1}$ and $\widetilde{V}_i$ for the recursive use of downsampling operations, applying element-wise addition directly on them will damage the boundary of the prediction region~\cite{huang2021fapn}. Thus, We leverage the deformable convolution to align $\widetilde{S}_{i+1}$ following the spatial location information of $\widetilde{V}_i$. Specifically, $\widetilde{V}_{i}$ and $\widetilde{S}_{i+1}$ are concatenated by the channel dimension, and $f_1(\cdot)$ is introduced to learn the offsets $\Delta_i$ which provides spatial differences between $\widetilde{V}_{i}$ and $\widetilde{S}_{i+1}$. Then, $f_2(\cdot)$ is introduced to align $\widetilde{S}_{i+1}$ according to transformation offsets of pixels:
\begin{equation}
	\Delta_i=f_1(Concat(\widetilde{V}_{i} ,\widetilde{S}_{i+1})),
\end{equation}
\begin{equation}
	\hat{S}_{i+1}=f_2(\widetilde{S}_{i+1}, \Delta_i),
\end{equation}
where $f_1(\cdot)$ and $f_2(\cdot)$ constitute the complete process of deformable convolution~\cite{Dai_2017_ICCV} with the kernel size of 3$\times$3. Ultimately, we apply element-wise addition on $\hat{S}_{i+1}$ and $\widetilde{V}_{i}$ to obtain higher-resolution segmentation map $S_i$, which provides rich semantic features for inpainting mapping.
\subsubsection{Attention-Based Filling Mapping.}
After obtaining the segmentation map $S_i$, we eliminate semantic information of pixels that are classified as the foreground and further introduce contextual attention to fill the missing region~\cite{zeng2019learning}. Specifically, we extract patches from $I_{i+1}$ and calculate the cosine similarity between patches inside and outside filled regions:
\begin{equation}
	S(I_{i+1}^m,I_{i+1}^n)=\left\langle\frac{I_{i+1}^m}{\left\|I_{i+1}^m\right\|_{2}}, \frac{I_{i+1}^n}{\left\|I_{i+1}^n\right\|_{2}}\right\rangle,
\end{equation}
where $I_{i+1}^m$, $I_{i+1}^n$ refers to the $m$-th external patch and the $n$-th internal patch extracted from $I_{i+1}$, respectively. For the $n$-th internal patch, the normalized context attention score with $m$-th external patch is obtained as follows:
\begin{equation}
	\alpha _{i+1}^{n,m}=\frac{\exp \left(S(I_{i+1}^m,I_{i+1}^n)\right)}{\sum_{m=1}^{M} \exp \left(S(I_{i+1}^m,I_{i+1}^n)\right)}.
\end{equation}
Then the eliminated holes in $S_i$ is filled with external context weighted by attention scores:
\begin{equation}
	I_{i}^{n}=\sum_{m=1}^{M} \alpha_{i+1}^{n,m} S_{i}^{m},
\end{equation}
where $I_i^{n}$ is the $n$-th patch to be filled into the eliminated region of $I_{i}$, and $S_i^{m}$ refers to the $m$-th external patch of $S_{i}$. Finally, we introduce three sets of hybrid dilated convolution (HDC)~\cite{YuKoltun2016} with expansion rates set as $\{1,2,5\}$ to aggregate multi-scale information and refine filling features. This mechanism enables $I_i$ to generate visually realistic semantic information with fine-grained details. 

\subsection{Loss Function}
Appropriate loss terms for segmentation and inpainting layers at each scale are designed to supervise multi-scale predictions. We adopt segmentation loss and inpainting loss to promote the fidelity of inpainted images.

\textbf{Segmentation Loss.} The segmentation loss is introduced to penalize deeply-supervised maps of segmentation, and we utilize binary cross entropy to progressively refine the predictions at each scale. 
\begin{equation}
	\mathcal{L}_{seg}=-\sum_{l=1}^{L-1} \sum_{p \in S} S(p) \log \left(h_1(\hat{S}^l)(p)\right),
\end{equation}
where $p$ is the pixel index, and $h_1(\cdot)$ upsamples $\hat{S}^l$ into the same size of $S$ .

\textbf{Inpainting Loss.} We utilize reconstruction loss $L_{rec}$ to compute the $L_1$ distance between each scale output and the corresponding ground-truth image:
\begin{equation}
	\mathcal{L}_{rec}=\sum_{l=1}^{L-1}\left\|{I}^l-h_2\left(\hat{I}^l\right)\right\|_1,
\end{equation}
where $h_2(\cdot)$ denotes a 1$\times$1 convolution which decodes $\hat{I}^l$ into an RGB image with the same size as $I^l$. To generate more realistic details, the extra adversarial loss with Patch-GAN~\cite{isola2017image} as the discriminator is also involved in the training process, 
\begin{equation}
	\mathcal{L}_{adv}=-\mathbb{E}\left[\log \left(1-\mathcal{D}_\omega(\hat{I})\right)\right]-\mathbb{E}\left[\log \mathcal{D}_\omega(I)\right],
\end{equation}
where $\mathcal{D}_\omega$ is the discriminator parameterized by $\omega$.

\textbf{Overall Training Loss.} The overall training loss function $\mathcal{L}_{tol}$ is the weighted sum of mentioned losses.
\begin{equation}
	\mathcal{L}_{tol}=\mathcal{L}_{seg}+\lambda_{rec}\mathcal{L}_{rec}+\lambda_{adv}\mathcal{L}_{adv},
\end{equation}
where the hyper-parameters $\lambda_{rec}$ and $\lambda_{adv}$ are empirically set as 5 and 20, respectively.
\section{Experiment}
\subsection{Implementation Details}
\textbf{Pre-training Strategy.} We adopt Swin-Transformer as the backbone to extract image features, and the SwinT layers employ the parameters SwinV2-G~\cite{liu2022swin}. Text embedding is initialized with official weights of ALBERT Base v2~\cite{DBLP} with 11M parameters and 12 layers. We introduce BIO annotation method~\cite{baldwin2009coding} to separately train the BiLSTM-CRF layer by first annotating partly data manually and utilizing word replacement templates to expend. To speed up the convergence of the model and improve the generalization ability, we further activate segmentation layers by downgrading the task as instance segmentation to strengthen the sensitivity to specific phrases with rich semantics. Broden+~\cite{xiao2018unified} is optimized as a training dataset and restrict referring expressions to the concepts like texture, material, and color. 

\textbf{Experimental Settings.} All the experiments are implemented on 4 V100 GPUs of 32G RAM. The input image is scaled to 480$\times$480, and the data augmentation strategy contains random cropping and flipping. The optimizer is AdamW~\cite{Loshchilov2017DecoupledWD} with $\beta_1=0.85$, $\beta_2=0.91$. The weight decay factor and the learning rate of the initial exponential decay are set to 0.005 and 0.0006, respectively, and the remaining parameters are randomly initialized. The model is trained for 76 rounds with the batch size set to 32 to achieve optimal performance.
\subsection{Metrics}
ROR task is a quintessential ill-posed task, and there are no applicable objective metrics due to the ambiguity of ground truth. The results are evaluated from two aspects including removal performance and computational overhead. We introduce PSNR, SSIM, LPIPS, and FID scores~\cite{zhang2018unreasonable} to comprehensively quantize the removal performance. We further measure computational overhead under the same hardware including parameters, floating point operations, and inference frames per second to report the effectiveness of real-time performance.
\subsection{Comparison with the State-of-the-arts}\label{5.3}
For quantitative analysis, we compare SAHM with diffusion-based models and two-stage methods which combine referring segmentation and image inpainting. We reproduce advanced models on ComCOCO, and all parameters are set consistent with original papers. Specifically, we retrain segmentation models on object annotations of the ComCOCO dataset to segment referred objects and then utilize inpainting models which are also separately retrained based on the filling information of ComCOCO to fill in the mask region. As manifested in~\cref{tab:example}, our model shows exponential improvement in terms of removal performance and computational overhead by a significant margin benefiting from the end-to-end structure. More ablative analysis is provided in \cref{appendixA}.

\begin{table*}[h]
	\setlength{\tabcolsep}{0.1pt}
	\caption{Comparison with state-of-the-arts on ComCOCO. We measure all the feasible combinations, and optimal results are demonstrated in the order of increasing PSNR.}
	\label{tab:example}
	\centering
	\begin{tabular}{cccccccc}
		\toprule
		\multirow{2}*{Model} & \multicolumn{4}{c}{Removal Performance}& \multicolumn{3}{c}{Computational Overhead}\\   \cmidrule(r){2-5} \cmidrule(r){6-8} 
		& PSNR$\uparrow$ & SSIM$\uparrow$ & LPIPS$\downarrow$ & FID$\downarrow$ &Params(M)&FLOPs(B)&FPS \\
		\midrule
		Instructpix2pix~\cite{brooks2023instructpix2pix}&11.23&0.356&0.674&38.74&534&69.26&0.18\\
		SmartEdit~\cite{huang2023smartedit}&12.26&0.447&0.616&35.90&749&70.88&0.07\\
		Inst-Inpaint~\cite{yildirim2023instinpaint}&15.92&0.527&0.510&32.67&942&94.53&0.10\\
		PolyFormer~\cite{Liu_2023_CVPR}+MI-GAN~\cite{sargsyan2023mi}&18.22&0.574&0.469&28.23&63&29.52&24\\
		 CGFormer~\cite{Tang_2023_CVPR}+SDXL~\cite{podell2023sdxl}&19.49&0.674&0.393&25.69&6788&474.05&0.14\\
		PolyFormer~\cite{Liu_2023_CVPR}+Repaint~\cite{lugmayr2022repaint}&19.64&0.683&0.331&25.00&712&89.08&0.12\\
		MCRES~\cite{xu2023meta}+LAMA~\cite{suvorov2021resolutionrobust} &20.08&0.748&0.411&30.64&236&56.23&16\\
		PolyFormer~\cite{Liu_2023_CVPR}+DiffIR~\cite{xia2023diffir}&20.12&0.769&0.305&22.62&97&21.89&0.74\\
		CRIS~\cite{Wang_2022_CVPR}+ZITS++~\cite{Dong_2022_CVPR} &20.97&0.773&0.343&26.08&102&35.46&27\\
		CRIS~\cite{Wang_2022_CVPR}+SCAT~\cite{zuo2023generative} &21.74 &0.802 &0.257 &27.51&142&40.69&20 \\
		\midrule
		SAHM & \textbf{23.29} & \textbf{0.831} &\textbf{0.204} &\textbf{22.08}&\textbf{42}&\textbf{10.42}&\textbf{136} \\
		\bottomrule
	\end{tabular}
\end{table*}

\begin{figure}[h]
	\centering
	\includegraphics[width=1.0\linewidth]{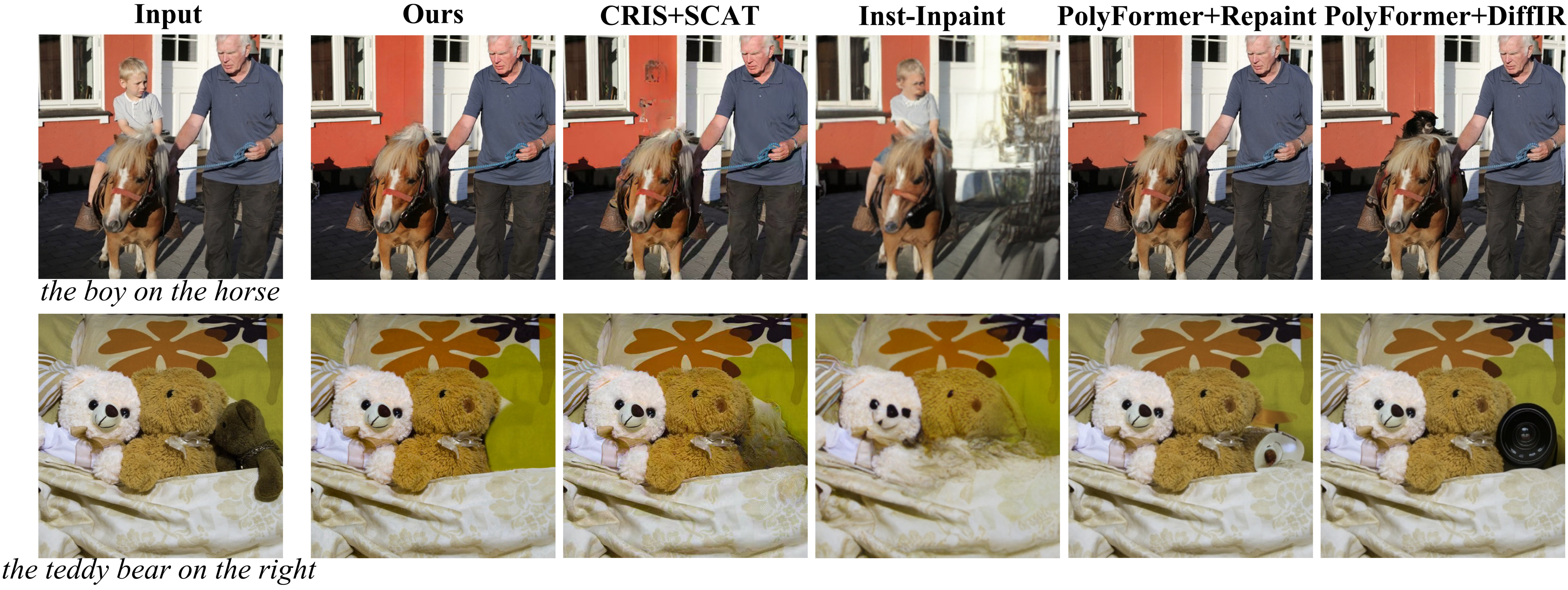}
	\caption{Exhibitions of removal results with different models.}
	\label{fig:onecol112}
	\vspace{-0.9cm}
\end{figure}

Diffusion-based inpainting models exhibit remarkable capability in generating satisfactory objects following textual descriptions, but their powerful generative capacities often lead to the generation of redundant object features by utilizing residual image semantics, as shown in \cref{fig:onecol112}. If the segmentaion model fails to precisely delineate the removal region, the remaining object features will exacerbate challenges faced by diffusion models. As for one-stage diffusion-based models, existing approaches show limitation in accurately locating the referred object in complex scenes while preserving irrelevant areas. Our further work will explore the collaborative potential between generative ability of diffusion models and our method.

\subsection{Ablation Study}
\textbf{Syntax-Aware Visual Attention.} We analyze the importance and stability of three head mechanisms in~\cref{tab:example2}. $L$-supervised attention as fundamental text embedding is an indispensable part of attention. We find that $L_{aw}$-supervised attention plays a greater role than $L_{iw}$ from a results-oriented perspective. This phenomenon can be interpreted that simple embedding $L$ fails to characterize $L_{aw}$ which is more informative than $L_{iw}$ and requires an additional specialized operation. We further add corresponding elements together in the weight matrix of three attention mechanisms and normalize them to contrast relative importance. The numerically reflected importance of $L$, $L_{aw}$ and $L_{iw}$ is 0.67, 0.24, and 0.19, respectively.

\begin{table}[h]
	\vspace{-0.2cm}
	\caption{Ablation results on ComCOCO with different semantic attention head settings.}
	\label{tab:example2}
	\centering
	\setlength{\tabcolsep}{5pt}
	\begin{tabular}{lcccccccc}
		\toprule
		\multicolumn{3}{c}{Settings} & \multicolumn{4}{c}{Metrics}\\  \cmidrule(r){1-3}  \cmidrule(r){4-7} 
		$L$      & $L_{aw}$       & $L_{iw}$       & PSNR$\uparrow$ & SSIM$\uparrow$ & LPIPS$\downarrow$ & FID$\downarrow$ \\
		\midrule
		\checkmark & & &21.76&0.714 &0.300 &23.13 \\
		\checkmark && \checkmark &22.19&0.711&0.317&22.85\\
		\checkmark & \checkmark&  &23.01&0.827&0.225&22.24\\
		\checkmark &\checkmark&\checkmark   &\textbf{23.29}&\textbf{0.831}&\textbf{0.204}&\textbf{22.08}\\
		\bottomrule
	\end{tabular}
\end{table}
\textbf{Feature Alignment.}  We further assess the effectiveness of feature alignment and hybrid dilated convolution. \Cref{tab:example3} indicates that two operations separately optimize the inpainting performance to some extent. The roughness of the location boundary potentially confuses the calculation of internal and external pixels, thus influencing subsequent inpainting. 

\begin{table}[h]
	\caption{Ablation results on ComCOCO concerning the feature alignment and hybrid dilated convolution in the upsampling process of hybrid mapping.}
	\label{tab:example3}
	\centering
	\setlength{\tabcolsep}{7.8pt}
	\begin{tabular}{l|cccccc}
		\toprule
		& PSNR$\uparrow$ & SSIM$\uparrow$ & LPIPS$\downarrow$ & FID$\downarrow$ \\
		\midrule
		w/o FA &22.00&0.763 &0.299 &24.06  \\
		w/ FA   &\textbf{23.29}&\textbf{0.831}&\textbf{0.204}&\textbf{22.08}\\
		\midrule
		w/o HDC &22.79&0.806 &0.330 &23.18 \\
		w/ HDC   &\textbf{23.29}&\textbf{0.831}&\textbf{0.204}&\textbf{22.08}\\
		\bottomrule
	\end{tabular}
	
\end{table}

\textbf{Loss Function.} We conduct experiments on different choices of loss functions in~\cref{tab:example6}. Due to the two-tier upsampling structure of the proposed model, the loss of segmentation layers significantly speeds up the training convergence and contributes to location accuracy.

\begin{table}[h]
	\caption{Ablation results on ComCOCO with different combinations of loss functions. The receptive field $N$ of Patch-GAN will be set very small if $L_{rec}$ is removed, thereby allowing the adversarial loss to guide the inpainting learning process. Training Time (TT) with normalization is introduced to measure the convergence rate of training under the same parameter settings.}
	\label{tab:example6}
	\centering
	\setlength{\tabcolsep}{4pt}
	\begin{tabular}{lcccccccc}
		\toprule
		\multicolumn{3}{c}{Loss} & \multicolumn{5}{c}{Metrics}\\  \cmidrule(r){1-3}  \cmidrule(r){4-8} 
		
		$\mathcal{L}_{seg}$      & $\mathcal{L}_{rec}$       & $\mathcal{L}_{adv}$      & PSNR$\uparrow$ & SSIM$\uparrow$ & LPIPS$\downarrow$ & FID$\downarrow$  & TT$\downarrow$ \\
		\midrule
		& \checkmark&\checkmark &19.32&0.656 &0.393 &25.77 &1\\
		\checkmark &\checkmark& &22.01&0.744&0.282&23.27&0.81\\
		\checkmark & & \checkmark &22.89&0.768&0.240&22.93&0.85\\
		\checkmark &\checkmark&\checkmark  &\textbf{23.29}&\textbf{0.831}&\textbf{0.204}&\textbf{22.08}&\textbf{0.79}\\
		\bottomrule
	\end{tabular}
\end{table}
\subsection{Qualitative Analysis}
\begin{figure}[h]
	\centering
	\includegraphics[width=1.0\linewidth]{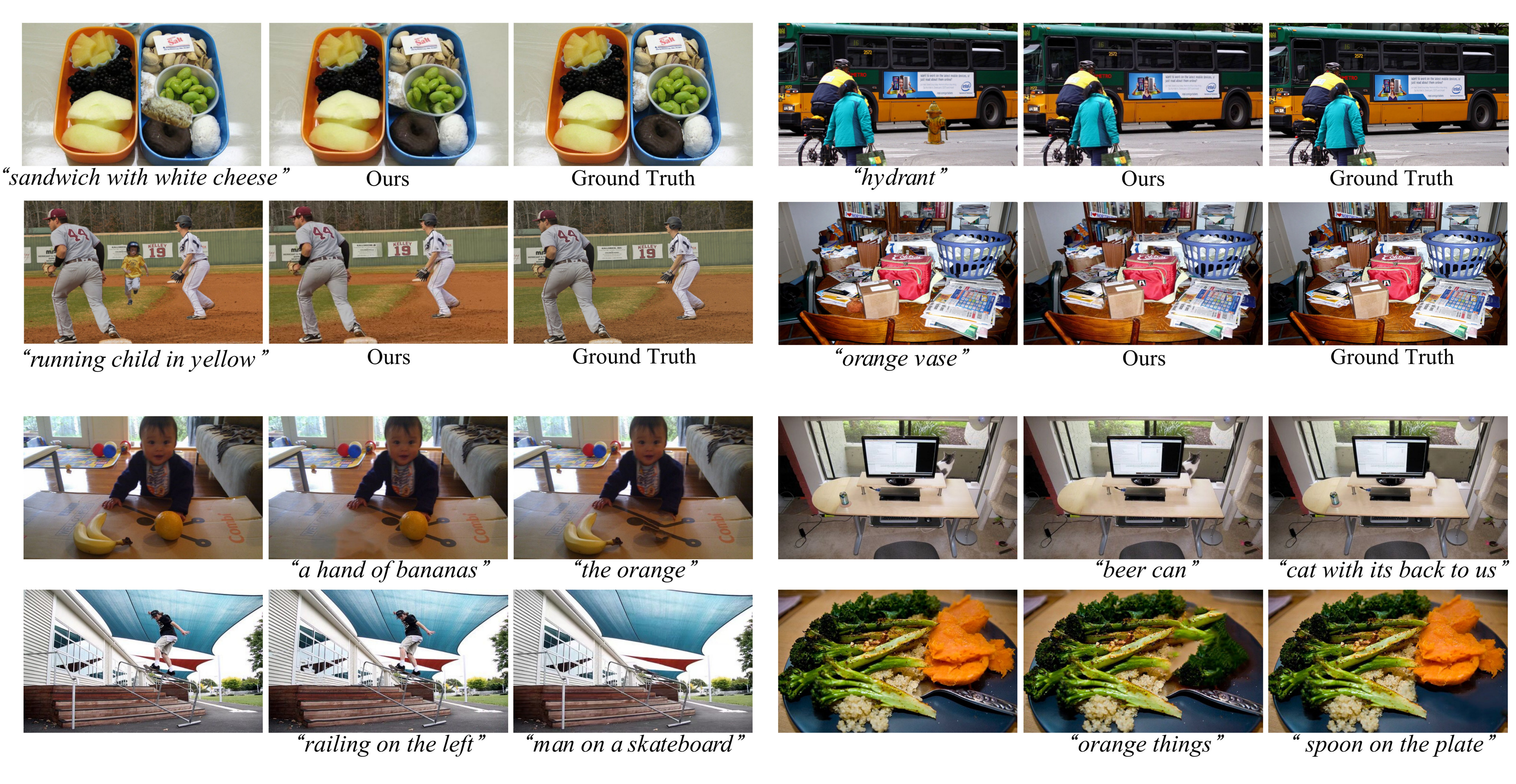}
	
	\caption{Visualization of referring object removal results. Images in the first two rows are derived from ComCOCO, and our result is visually compared with the ground truth. The last two rows show realistic images with no gold standard. The first one in each group is the original image, and the next two images are our elimination results under the guidance of expressions.}
	\label{fig:onecol10}
\end{figure}

\Cref{fig:onecol10} visualizes partial examples of our results in terms of synthetic scenes and realistic natural scenes to demonstrate the generalization of the proposed model and the practicability of our dataset. The algorithm shows remarkable ability to locate numerous kinds of objects in complex semantic scenarios and fill vacancies creatively according to the surrounding semantic information. Real-world applications and more qualitative analysis are provided in \cref{appendixC}.

\section{Conclusion}
In this paper, we rethink referring object removal task with a new perspective. The common challenge in this field is the collection of real-world local changes, and ComCOCO serves as the benchmark dataset for researchers to apply downstream transformations and it can be applied to related tasks. We also explore the possibility of joint learning of multiple tasks, and it is instructive for follow-up research areas. 
\clearpage
\bibliographystyle{splncs04}
\bibliography{main}

\begin{thebibliography}{10}
\providecommand{\url}[1]{\texttt{#1}}
\providecommand{\urlprefix}{URL }
\providecommand{\doi}[1]{https://doi.org/#1}

\bibitem{baldwin2009coding}
Baldwin, B.: Coding chunkers as taggers: Io, bio, bmewo, and bmewo+. LingPipe
  Blog, October  \textbf{14} (2009)

\bibitem{brooks2023instructpix2pix}
Brooks, T., Holynski, A., Efros, A.A.: Instructpix2pix: Learning to follow
  image editing instructions (2023)

\bibitem{cao2022learning}
Cao, C., Dong, Q., Fu, Y.: Learning prior feature and attention enhanced image
  inpainting (2022)

\bibitem{Dai_2017_ICCV}
Dai, J., Qi, H., Xiong, Y., Li, Y., Zhang, G., Hu, H., Wei, Y.: Deformable
  convolutional networks. In: Proceedings of the IEEE International Conference
  on Computer Vision (ICCV) (Oct 2017)

\bibitem{Ding_2021_ICCV}
Ding, H., Liu, C., Wang, S., Jiang, X.: Vision-language transformer and query
  generation for referring segmentation. In: Proceedings of the IEEE/CVF
  International Conference on Computer Vision (ICCV). pp. 16321--16330 (October
  2021)

\bibitem{Dong_2022_CVPR}
Dong, Q., Cao, C., Fu, Y.: Incremental transformer structure enhanced image
  inpainting with masking positional encoding. In: Proceedings of the IEEE/CVF
  Conference on Computer Vision and Pattern Recognition (CVPR). pp.
  11358--11368 (June 2022)

\bibitem{Feng_2021_CVPR}
Feng, G., Hu, Z., Zhang, L., Lu, H.: Encoder fusion network with co-attention
  embedding for referring image segmentation. In: Proceedings of the IEEE/CVF
  Conference on Computer Vision and Pattern Recognition (CVPR). pp.
  15506--15515 (June 2021)

\bibitem{Guo_2021_ICCV}
Guo, X., Yang, H., Huang, D.: Image inpainting via conditional texture and
  structure dual generation. In: Proceedings of the IEEE/CVF International
  Conference on Computer Vision (ICCV). pp. 14134--14143 (October 2021)

\bibitem{guo2021image}
Guo, X., Yang, H., Huang, D.: Image inpainting via conditional texture and
  structure dual generation. In: Proceedings of the IEEE/CVF International
  Conference on Computer Vision. pp. 14134--14143 (2021)

\bibitem{huang2021fapn}
Huang, S., Lu, Z., Cheng, R., He, C.: Fapn: Feature-aligned pyramid network for
  dense image prediction. In: Proceedings of the IEEE/CVF International
  Conference on Computer Vision. pp. 864--873 (2021)

\bibitem{huang2023smartedit}
Huang, Y., Xie, L., Wang, X., Yuan, Z., Cun, X., Ge, Y., Zhou, J., Dong, C.,
  Huang, R., Zhang, R., Shan, Y.: Smartedit: Exploring complex
  instruction-based image editing with multimodal large language models (2023)

\bibitem{isola2017image}
Isola, P., Zhu, J.Y., Zhou, T., Efros, A.A.: Image-to-image translation with
  conditional adversarial networks. In: Proceedings of the IEEE conference on
  computer vision and pattern recognition. pp. 1125--1134 (2017)

\bibitem{Harmonizer}
Ke, Z., Sun, C., Zhu, L., Xu, K., Lau, R.W.: Harmonizer: Learning to perform
  white-box image and video harmonization. In: European Conference on Computer
  Vision (ECCV) (2022)

\bibitem{Kim2017}
Kim, J.H., On, K.W., Lim, W., Kim, J., Ha, J.W., Zhang, B.T.: {Hadamard Product
  for Low-rank Bilinear Pooling}. In: The 5th International Conference on
  Learning Representations (2017)

\bibitem{kim2022restr}
Kim, N., Kim, D., Lan, C., Zeng, W., Kwak, S.: Restr: Convolution-free
  referring image segmentation using transformers. In: Proceedings of the
  IEEE/CVF Conference on Computer Vision and Pattern Recognition. pp.
  18145--18154 (2022)

\bibitem{lample-etal-2016-neural}
Lample, G., Ballesteros, M., Subramanian, S., Kawakami, K., Dyer, C.: Neural
  architectures for named entity recognition. In: Proceedings of the 2016
  Conference of the North {A}merican Chapter of the Association for
  Computational Linguistics: Human Language Technologies. pp. 260--270.
  Association for Computational Linguistics, San Diego, California (Jun 2016)

\bibitem{DBLP}
Lan, Z., Chen, M., Goodman, S., Gimpel, K., Sharma, P., Soricut, R.: {ALBERT:}
  {A} lite {BERT} for self-supervised learning of language representations. In:
  8th International Conference on Learning Representations, {ICLR} 2020, Addis
  Ababa, Ethiopia, April 26-30, 2020. OpenReview.net (2020)

\bibitem{lin2014microsoft}
Lin, T.Y., Maire, M., Belongie, S., Hays, J., Perona, P., Ramanan, D.,
  Doll{\'a}r, P., Zitnick, C.L.: Microsoft coco: Common objects in context. In:
  European conference on computer vision. pp. 740--755. Springer (2014)

\bibitem{Liu_2023_CVPR}
Liu, J., Ding, H., Cai, Z., Zhang, Y., Satzoda, R.K., Mahadevan, V., Manmatha,
  R.: Polyformer: Referring image segmentation as sequential polygon
  generation. In: Proceedings of the IEEE/CVF Conference on Computer Vision and
  Pattern Recognition (CVPR). pp. 18653--18663 (June 2023)

\bibitem{liu2022reduce}
Liu, Q., Tan, Z., Chen, D., Chu, Q., Dai, X., Chen, Y., Liu, M., Yuan, L., Yu,
  N.: Reduce information loss in transformers for pluralistic image inpainting.
  In: Proceedings of the IEEE/CVF Conference on Computer Vision and Pattern
  Recognition. pp. 11347--11357 (2022)

\bibitem{liu2022swin}
Liu, Z., Hu, H., Lin, Y., Yao, Z., Xie, Z., Wei, Y., Ning, J., Cao, Y., Zhang,
  Z., Dong, L., et~al.: Swin transformer v2: Scaling up capacity and
  resolution. In: Proceedings of the IEEE/CVF Conference on Computer Vision and
  Pattern Recognition. pp. 12009--12019 (2022)

\bibitem{Liu_2021_ICCV}
Liu, Z., Lin, Y., Cao, Y., Hu, H., Wei, Y., Zhang, Z., Lin, S., Guo, B.: Swin
  transformer: Hierarchical vision transformer using shifted windows. In:
  Proceedings of the IEEE/CVF International Conference on Computer Vision
  (ICCV). pp. 10012--10022 (October 2021)

\bibitem{liu2021swin}
Liu, Z., Lin, Y., Cao, Y., Hu, H., Wei, Y., Zhang, Z., Lin, S., Guo, B.: Swin
  transformer: Hierarchical vision transformer using shifted windows. In:
  Proceedings of the IEEE/CVF International Conference on Computer Vision. pp.
  10012--10022 (2021)

\bibitem{liu2015deep}
Liu, Z., Luo, P., Wang, X., Tang, X.: Deep learning face attributes in the
  wild. In: Proceedings of the IEEE international conference on computer
  vision. pp. 3730--3738 (2015)

\bibitem{Loshchilov2017DecoupledWD}
Loshchilov, I., Hutter, F.: Decoupled weight decay regularization. In:
  International Conference on Learning Representations (2017)

\bibitem{lugmayr2022repaint}
Lugmayr, A., Danelljan, M., Romero, A., Yu, F., Timofte, R., Van~Gool, L.:
  Repaint: Inpainting using denoising diffusion probabilistic models. In:
  Proceedings of the IEEE/CVF Conference on Computer Vision and Pattern
  Recognition. pp. 11461--11471 (2022)

\bibitem{Luo_2020_CVPR}
Luo, G., Zhou, Y., Sun, X., Cao, L., Wu, C., Deng, C., Ji, R.: Multi-task
  collaborative network for joint referring expression comprehension and
  segmentation. In: Proceedings of the IEEE/CVF Conference on Computer Vision
  and Pattern Recognition (CVPR) (June 2020)

\bibitem{mao2016generation}
Mao, J., Huang, J., Toshev, A., Camburu, O., Yuille, A.L., Murphy, K.:
  Generation and comprehension of unambiguous object descriptions. In:
  Proceedings of the IEEE conference on computer vision and pattern
  recognition. pp. 11--20 (2016)

\bibitem{2013Using}
Niwattanakul, S., Singthongchai, J., Naenudorn, E., Wanapu, S.: Using of
  jaccard coefficient for keywords similarity. In: Iaeng International
  Conference on Internet Computing and Web Services (2013)

\bibitem{podell2023sdxl}
Podell, D., English, Z., Lacey, K., Blattmann, A., Dockhorn, T., M{\"u}ller,
  J., Penna, J., Rombach, R.: Sdxl: Improving latent diffusion models for
  high-resolution image synthesis. arXiv preprint arXiv:2307.01952  (2023)

\bibitem{sargsyan2023mi}
Sargsyan, A., Navasardyan, S., Xu, X., Shi, H.: Mi-gan: A simple baseline for
  image inpainting on mobile devices. In: Proceedings of the IEEE/CVF
  International Conference on Computer Vision. pp. 7335--7345 (2023)

\bibitem{song2018contextual}
Song, Y., Yang, C., Lin, Z., Liu, X., Huang, Q., Li, H., Kuo, C.C.J.:
  Contextual-based image inpainting: Infer, match, and translate. In:
  Proceedings of the European Conference on Computer Vision (ECCV). pp. 3--19
  (2018)

\bibitem{suvorov2021resolutionrobust}
Suvorov, R., Logacheva, E., Mashikhin, A., Remizova, A., Ashukha, A.,
  Silvestrov, A., Kong, N., Goka, H., Park, K., Lempitsky, V.:
  Resolution-robust large mask inpainting with fourier convolutions (2021)

\bibitem{Tang_2023_CVPR}
Tang, J., Zheng, G., Shi, C., Yang, S.: xu2023bridging. In: Proceedings of the
  IEEE/CVF Conference on Computer Vision and Pattern Recognition (CVPR). pp.
  23570--23580 (June 2023)

\bibitem{2018Training}
Tremblay, J., Prakash, A., Acuna, D., Brophy, M., Jampani, V., Anil, C., To,
  T., Cameracci, E., Boochoon, S., Birchfield, S.: Training deep networks with
  synthetic data: Bridging the reality gap by domain randomization. 2018
  IEEE/CVF Conference on Computer Vision and Pattern Recognition Workshops
  (CVPRW)  (2018)

\bibitem{tylevcek2013spatial}
Tyle{\v{c}}ek, R., {\v{S}}{\'a}ra, R.: Spatial pattern templates for
  recognition of objects with regular structure. In: German conference on
  pattern recognition. pp. 364--374. Springer (2013)

\bibitem{NIPS2017_3f5ee243}
Vaswani, A., Shazeer, N., Parmar, N., Uszkoreit, J., Jones, L., Gomez, A.N.,
  Kaiser, L.u., Polosukhin, I.: Attention is all you need. In: Guyon, I.,
  Luxburg, U.V., Bengio, S., Wallach, H., Fergus, R., Vishwanathan, S.,
  Garnett, R. (eds.) Advances in Neural Information Processing Systems.
  vol.~30. Curran Associates, Inc. (2017)

\bibitem{wang2021image}
Wang, T., Ouyang, H., Chen, Q.: Image inpainting with external-internal
  learning and monochromic bottleneck. In: Proceedings of the IEEE/CVF
  Conference on Computer Vision and Pattern Recognition. pp. 5120--5129 (2021)

\bibitem{Wang_2022_CVPR}
Wang, Z., Lu, Y., Li, Q., Tao, X., Guo, Y., Gong, M., Liu, T.: Cris:
  Clip-driven referring image segmentation. In: Proceedings of the IEEE/CVF
  Conference on Computer Vision and Pattern Recognition (CVPR). pp.
  11686--11695 (June 2022)

\bibitem{wu2020phrasecut}
Wu, C., Lin, Z., Cohen, S., Bui, T., Maji, S.: Phrasecut: Language-based image
  segmentation in the wild. In: Proceedings of the IEEE/CVF Conference on
  Computer Vision and Pattern Recognition. pp. 10216--10225 (2020)

\bibitem{xia2023diffir}
Xia, B., Zhang, Y., Wang, S., Wang, Y., Wu, X., Tian, Y., Yang, W., Van~Gool,
  L.: Diffir: Efficient diffusion model for image restoration. arXiv preprint
  arXiv:2303.09472  (2023)

\bibitem{xiao2018unified}
Xiao, T., Liu, Y., Zhou, B., Jiang, Y., Sun, J.: Unified perceptual parsing for
  scene understanding. In: Proceedings of the European conference on computer
  vision (ECCV). pp. 418--434 (2018)

\bibitem{pmlr-v37-xuc15}
Xu, K., Ba, J., Kiros, R., Cho, K., Courville, A., Salakhudinov, R., Zemel, R.,
  Bengio, Y.: Show, attend and tell: Neural image caption generation with
  visual attention. In: Bach, F., Blei, D. (eds.) Proceedings of the 32nd
  International Conference on Machine Learning. Proceedings of Machine Learning
  Research, vol.~37, pp. 2048--2057. PMLR, Lille, France (07--09 Jul 2015)

\bibitem{xu2023meta}
Xu, L., Huang, M.H., Shang, X., Yuan, Z., Sun, Y., Liu, J.: Meta compositional
  referring expression segmentation. In: Proceedings of the IEEE/CVF Conference
  on Computer Vision and Pattern Recognition (CVPR) (June 2023)

\bibitem{yang2023semanticsaware}
Yang, Z., Wang, J., Tang, Y., Chen, K., Zhao, H., Torr, P.H.S.: Semantics-aware
  dynamic localization and refinement for referring image segmentation (2023)

\bibitem{Yang_2022_CVPR}
Yang, Z., Wang, J., Tang, Y., Chen, K., Zhao, H., Torr, P.H.: Lavt:
  Language-aware vision transformer for referring image segmentation. In:
  Proceedings of the IEEE/CVF Conference on Computer Vision and Pattern
  Recognition (CVPR). pp. 18155--18165 (June 2022)

\bibitem{yildirim2023instinpaint}
Yildirim, A.B., Baday, V., Erdem, E., Erdem, A., Dundar, A.: Inst-inpaint:
  Instructing to remove objects with diffusion models (2023)

\bibitem{YuKoltun2016}
Yu, F., Koltun, V.: Multi-scale context aggregation by dilated convolutions.
  In: ICLR (2016)

\bibitem{2016Modeling}
Yu, L., Poirson, P., Yang, S., Berg, A.C., Berg, T.L.: Modeling context in
  referring expressions. In: Springer International Publishing (2016)

\bibitem{zeng2019learning}
Zeng, Y., Fu, J., Chao, H., Guo, B.: Learning pyramid-context encoder network
  for high-quality image inpainting. In: Proceedings of the IEEE/CVF Conference
  on Computer Vision and Pattern Recognition. pp. 1486--1494 (2019)

\bibitem{zeng2021cr}
Zeng, Y., Lin, Z., Lu, H., Patel, V.M.: Cr-fill: Generative image inpainting
  with auxiliary contextual reconstruction. In: Proceedings of the IEEE/CVF
  International Conference on Computer Vision. pp. 14164--14173 (2021)

\bibitem{zhang2018unreasonable}
Zhang, R., Isola, P., Efros, A.A., Shechtman, E., Wang, O.: The unreasonable
  effectiveness of deep features as a perceptual metric. In: Proceedings of the
  IEEE conference on computer vision and pattern recognition. pp. 586--595
  (2018)

\bibitem{7968387}
Zhou, B., Lapedriza, A., Khosla, A., Oliva, A., Torralba, A.: Places: A 10
  million image database for scene recognition. IEEE Transactions on Pattern
  Analysis and Machine Intelligence  \textbf{40}(6),  1452--1464 (2018)

\bibitem{10.1007/978-3-031-19790-1_23}
Zhou, S., Liu, L., Niu, L., Zhang, L.: Learning object placement via dual-path
  graph completion. In: Avidan, S., Brostow, G., Ciss{\'e}, M., Farinella,
  G.M., Hassner, T. (eds.) Computer Vision -- ECCV 2022. pp. 373--389. Springer
  Nature Switzerland, Cham (2022)

\bibitem{zuo2023generative}
Zuo, Z., Zhao, L., Li, A., Wang, Z., Zhang, Z., Chen, J., Xing, W., Lu, D.:
  Generative image inpainting with segmentation confusion adversarial training
  and contrastive learning (2023)

\end{thebibliography}
\clearpage
\begin{appendix}
	\section{Ablative Analysis of Model Architectures}\label{appendixA}
	\Cref{5.3} compares our model with state-of-the-arts to experimentally confirm the considerable advancement. We further conduct a series of experiments on public datasets with similar architectural choices to confirm the necessity of the end-to-end structure.
	\subsection{Segmentation Module}
	Following retaining the original structure of the model, we separate the inpainting layer in the upsampling stage and reconstruct the feature map to output the segmentation results.     Intersection over Union (IoU) is introduced as the metric to comprehensively evaluate the capability of pixel-level mask generation and object location. As demonstrated in~\cref{tab:example10}, the segmentation architecture of SAHM fails to surpass all the SOTAs. We deliberate on advanced model structure for reference and endeavor to optimize performance by appending optimization modules (\cref{4.1} and \cref{4.2.1}). Nevertheless, transcending state-of-the-arts in segmentation is arduous in terms of distinct modeling purposes, and the considerable advantage of our model on the ROR task is not derived from fine-grained segmentation.     The performance of segmentation models in ComCOCO (horizontally compared with public datasets and vertically compared with SOTAs) indicates the validity of the synthetic dataset to some extent.
	
		\begin{table}[h]
		\caption{Comparison of segmentation performance with state-of-the-arts on public datasets and our proposed ComCOCO.}
		\label{tab:example10}
		\setlength{\tabcolsep}{0.8pt}
		\centering
		\begin{tabular}{ccccccccccc}
			\toprule
			\multirow{2}{*}{Model} &\multirow{2}{*}{Params}& \multicolumn{3}{c}{RefCOCO~\cite{2016Modeling}}& \multicolumn{3}{c}{RefCOCO+~\cite{2016Modeling}}& \multicolumn{1}{c}{G-Ref~\cite{mao2016generation}}&\multicolumn{1}{c}{ComCOCO}\\  \cmidrule(r){3-5}  \cmidrule(r){6-8} \cmidrule(r){9-9}  \cmidrule(r){10-10} 
			
			&(M)&val    &testA &testB & val & testA &testB &val &test\\
			\midrule
			EFN~\cite{Feng_2021_CVPR}&126.89&62.76&65.69&59.67&51.50&55.24&43.01&51.93&54.73\\
			ReSTR~\cite{kim2022restr} &86.19&67.22&69.30&64.45&55.78&60.44&48.27&54.48&62.23\\
			CRIS~\cite{Wang_2022_CVPR} &57.31&70.47&73.18&66.10&62.27&68.05&53.68&59.87&65.25\\
			LAVT~\cite{Yang_2022_CVPR}&129.36&72.73&75.82&68.79&62.14&68.38&55.10&60.50&64.99\\
			CGFormer~\cite{Tang_2023_CVPR}&63.67&74.75&77.30&70.64&64.54 &71.00 &57.14 &\textbf{64.68 }&68.55\\
		CRES~\cite{xu2023meta}&143.05&74.92&76.98&70.84&64.32&69.68 &56.64 &-&67.24\\
			PolyFormer~\cite{Liu_2023_CVPR}&95.12&\textbf{75.96}&\textbf{78.29}&\textbf{73.25}&\textbf{69.33}&\textbf{74.56} &\textbf{ 61.87 }&-&\textbf{69.01}\\
			\midrule
			SAHM$_{Seg}$&46.44&73.47&76.29&69.13&66.48&70.03&56.31&63.85&68.56\\
			\bottomrule
		\end{tabular}
	\end{table}
	\subsection{Inpainting Module}
	In SAHM, the missing region is filled with affinity learned from high-level semantic features in the upsampling stage to accomplish image inpainting.   We reserve inpainting layers and restructure the encoding of image feature maps with the corresponding downsampling methods based on the backbone (\ie Swin-Transformer). The modified SAHM is compared with state-of-the-art approaches in~\cref{tab:example8}.   We experiment on two datasets: Places2~\cite{7968387} with over 8,000,000 images from over 365 natural scenes to evaluate the inpainting performance on random holes with various scales, and ComCOCO with object holes to place extra emphasis on complete-block filling. Although it is not a necessity to introduce our proposed dataset here since its foreground information is thoroughly removed which is equivalent to adding an arbitrary hole to an image, the comparison of inpainting metrics with Places2 on the same approach verifies the difficulty coefficient of the synthetic ComCOCO. Our model possesses similar performance in a single inpainting task with fewer parameters. SAHM centers on the common end-to-end baseline to solve the ROR task, and the superior capability barely owes to the inpainting structure.
	
		\begin{table}[h]
		\caption{Comparison of inpainting performance with state-of-the-arts on Places2 and our proposed ComCOCO. }
		\label{tab:example8}
		\setlength{\tabcolsep}{5pt}
		\centering
		\begin{tabular}{cccccccccc}
			\toprule
			\multirow{2}{*}{Model}  &\multirow{2}{*}{Params}& \multicolumn{3}{c}{Places2~\cite{7968387}}& \multicolumn{3}{c}{ComCOCO}& \\  \cmidrule(r){3-5}  \cmidrule(r){6-8} 
			&(M)&PSNR$\uparrow$    &SSIM$\uparrow$ &FID$\downarrow$ &PSNR$\uparrow$    &SSIM$\uparrow$ &FID$\downarrow$ \\
			\midrule
			SDXL~\cite{podell2023sdxl}&6600&23.67&0.738&15.41&20.38&0.683&23.34\\
			PUT~\cite{liu2022reduce}&29&24.49&0.806&22.12&27.91&0.807&18.62\\
			RePaint~\cite{lugmayr2022repaint}&607&24.97&0.854&4.07&23.56&0.721&20.73\\
			SCAT~\cite{zuo2023generative}&48&25.48&0.840&6.98&\textbf{31.11}&0.859&13.33\\
			DiffIR~\cite{xia2023diffir}&27&\textbf{26.23}&\textbf{0.891}&\textbf{1.98}&24.71&0.735&12.44\\
			\midrule
			SAHM$_{Inp}$&6& 24.15 &0.792&3.64&29.11&\textbf{0.871}&\textbf{10.60}\\
			\bottomrule
		\end{tabular}
	\end{table}
	
	\subsection{Advantages of the End-to-end Structure}
	Intuitively, the segmentation veracity correlates closely with inpainting performance considering that the inpainting process is conducted in the mask region.  Nonetheless, there is no necessary causal connection, which is demonstrated by comprehensive experiments.  We concatenate multiple segmentation models with an off-the-shelf inpainting model to solve the ROR task in a two-stage modality.  As manifested in~\cref{tab:example11}, the performance of three referring segmentation models is enhanced sequentially, and the filling results do not follow a corresponding pattern of change when the same inpainting model is applied to their removing regions.
	
	Sufficient exemplifications refute the common misconception that the exactness of location consequentially contributes to the superior performance of inpainting.   The deviation of segmentation retains the residue of foreground objects and eliminates a fraction of background pixels, which introduces unforeseeable uncertainty when learning semantic features around missing regions.   Alongside this, considering the connatural limitation of roughly delineating the boundary of specific objects in the ground truth of segmentation masks, the two-stage approach will generate an accumulative error.   In the final analysis, the outperformance of SAHM benefits principally from the end-to-end structure and joint training strategy.
	
	\begin{table}[h]
		\caption{Comparison of segmentation models concatenated with the same inpainting model on ComCOCO. The best and second best results are highlighted in \textbf{boldfaced} and \underline{underlined}, respectively.}
		\label{tab:example11}
		\setlength{\tabcolsep}{3pt}
		\centering
		\begin{tabular}{cccccccc}
			\toprule
			\multicolumn{2}{c}{Model} & \multicolumn{2}{c}{Metrics$_{Seg}$}&\multicolumn{3}{c}{Metrics$_{Inp}$} &\\  \cmidrule(r){1-2} \cmidrule(r){3-4}  \cmidrule(r){5-8} 
			
			Seg &Inp  &IOU &Pr@0.7&PSNR &SSIM&FID \\
			\midrule
			RESTR\cite{kim2022restr}&	\multirow{3}{*}{ZITS\cite{Dong_2022_CVPR}}& 62.23 & 56.11  &16.07&0.482&\textbf{29.12}\\
			LAVT\cite{Yang_2022_CVPR} & &\underline{64.99}& \underline{58.90} & \textbf{17.74}&\textbf{0.580}&\underline{29.57}\\
			CRIS\cite{Wang_2022_CVPR}& &\textbf{65.25}&\textbf{59.23} &\underline{16.91}&\underline{0.548}&30.64\\
			\bottomrule
		\end{tabular}
		
	\end{table}

	\section{Elaboration on ComCOCO}\label{appendixB}
	
\begin{figure*}[h]
		\centering
		\includegraphics[width=1.0\linewidth]{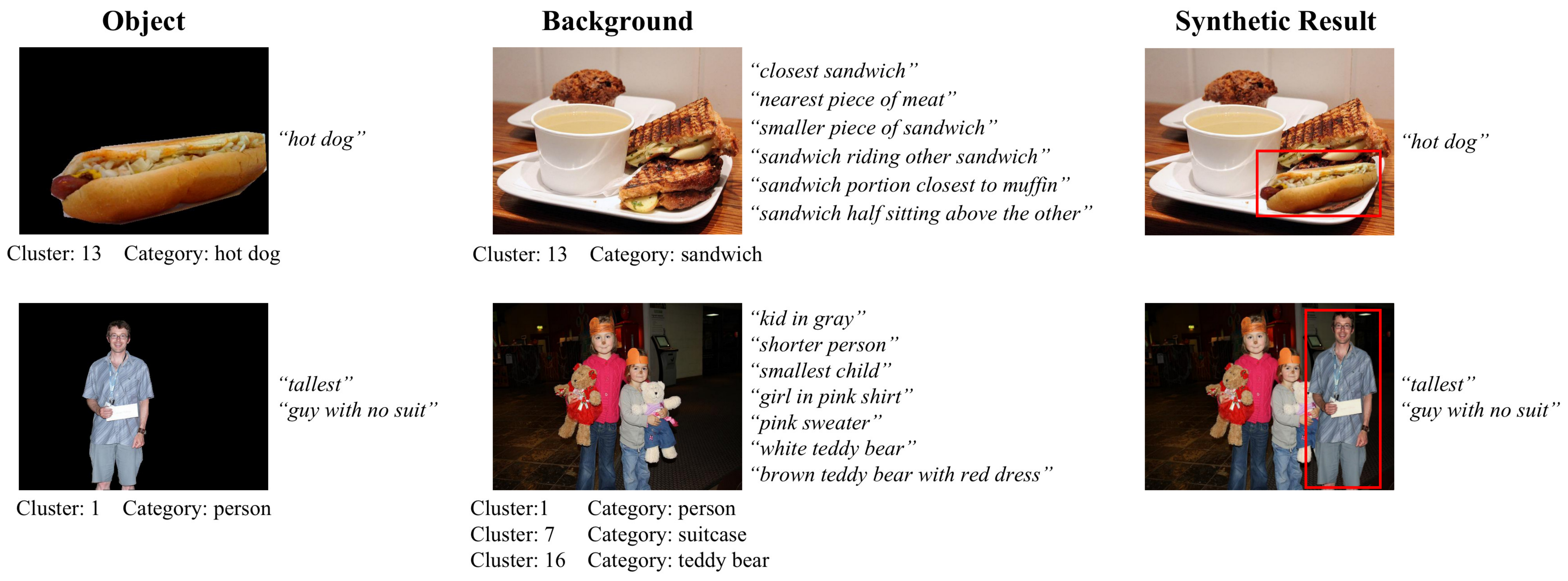}
		
		\caption{Exhibition of selected object-background pairs in the construction process. The first column is the object with descriptive sentences. The second column is selected background with descriptions of all the objects in the scene, and the sentence similarity is introduced to estimate whether the added object has similar attributes to the objects in the background. The last column is the synthetic result. }
		\label{fig:onecol12}
	\end{figure*}

\begin{figure}[h]
		\begin{subfigure}{0.4\linewidth}
			\begin{minipage}[t]{0.5\linewidth}
				\includegraphics[width=1\linewidth]{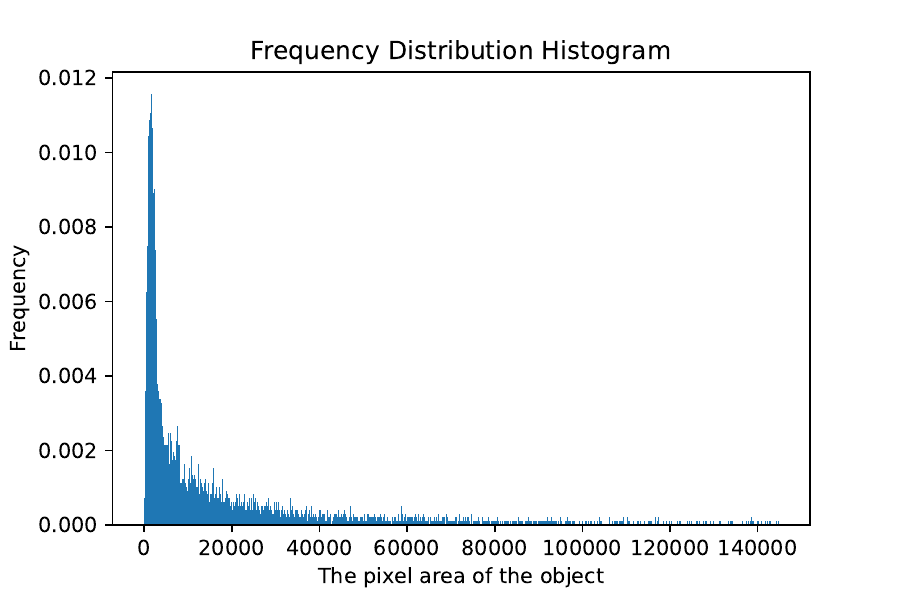}
				\caption{Plot A}
				\label{0}
				\includegraphics[width=2\linewidth]{test11.eps}
				\caption{{Plot B}}
				\label{-a}
			\end{minipage}%
			\includegraphics[width=0.5\linewidth]{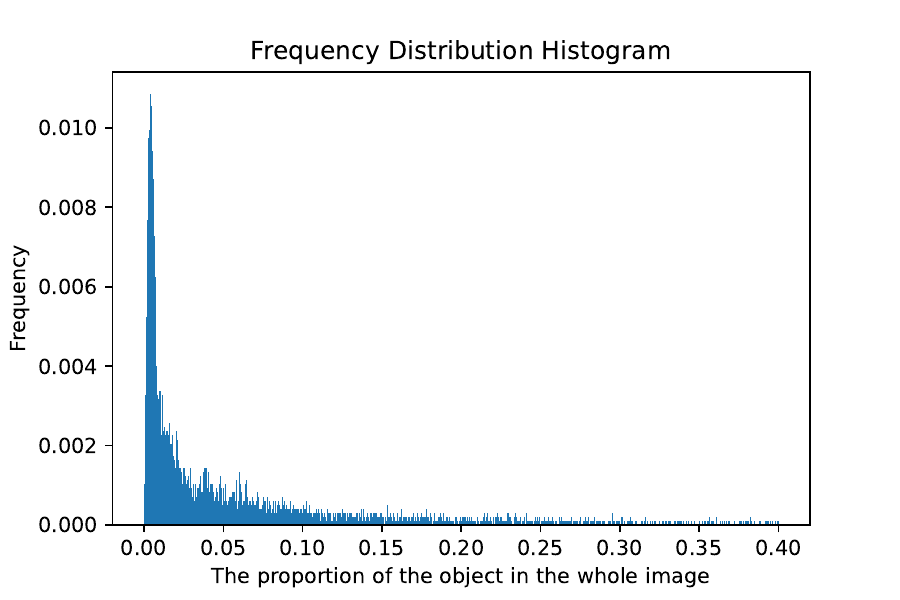}
			\label{fig:short-4}
		\end{subfigure}
		\begin{subfigure}{0.6\linewidth}
			\includegraphics[width=1\linewidth]{test3.eps}
			\label{fig:short-c}
			\caption{\centering{Plot C} }
		\end{subfigure}
		\caption{{Analysis of the ComCOCO dataset. Plot A shows the frequency distribution of the pixel area and proportion of objects. Plot B shows the heat map of object distribution in images, where all the images are resized to 320$\times$200 and the value in each coordinate refers to the frequency of object masks appearing in the corresponding pixel. Plot C contrasts the number of instances per cluster and contained categories.}}
		\label{fig:short}
	\end{figure}

\begin{figure*}[h]
		\centering
		\includegraphics[width=0.97\linewidth]{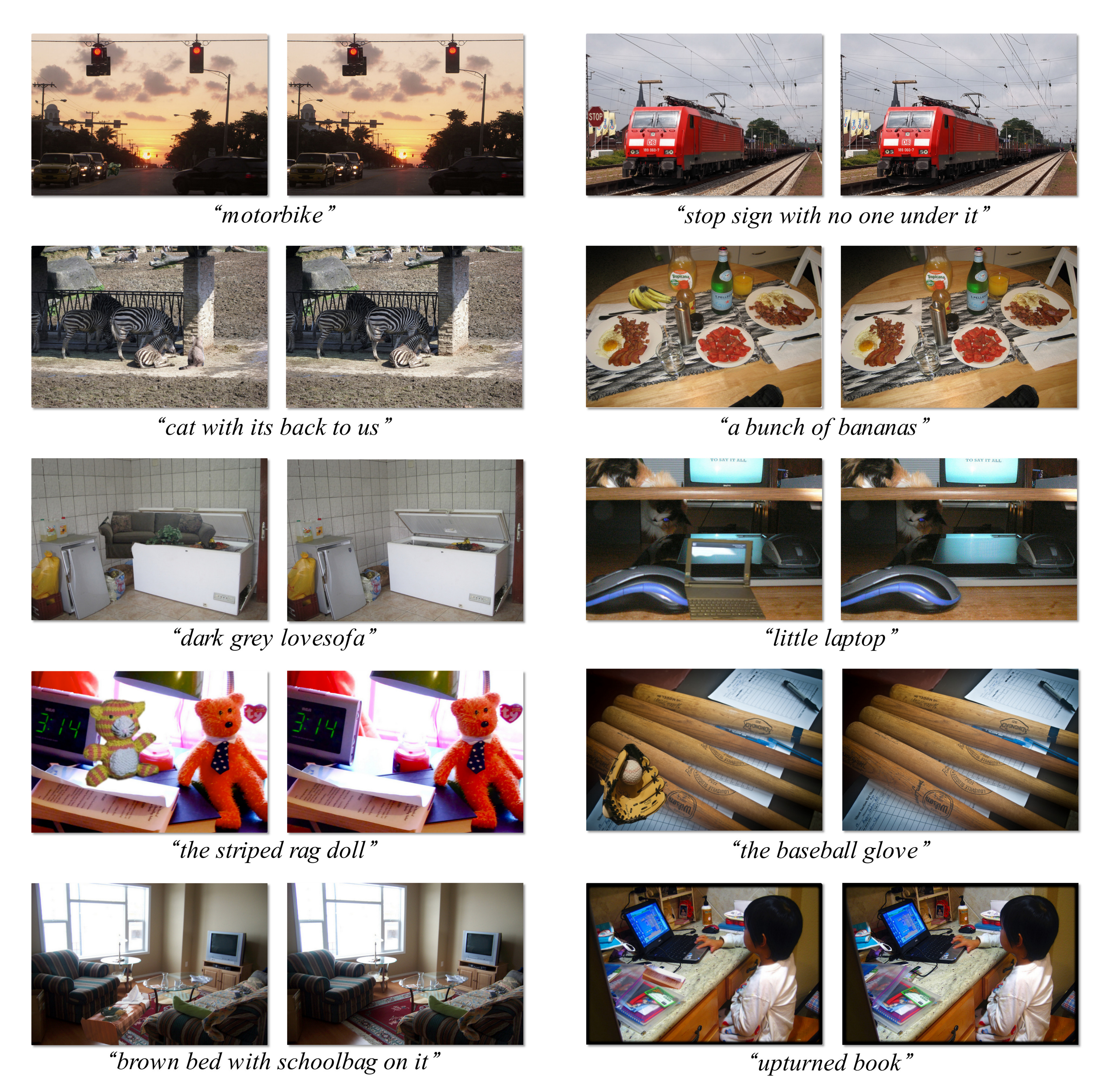}
		\caption{Exhibition of images in ComCOCO. The first image in each pair is the input, and the second one is the ground truth.}
		\label{fig:onecol11}
	\end{figure*}

	\textbf{Realism of ComCOCO.} We adopt a combination of automated construction processes and meticulous manual inspection, aimed to mitigate the introduction of artifacts that could potentially be exploited by complex deep neural networks. As for Step 1 (Scene Matching),~\cref{tab:example9} enumerates 17 clusters for 80 object categories, and~\cref{fig:onecol12} demonstrates the reliability of the construction process. Manual inspection is performed separately after Step 2 (Object Placement) and Step 3 (Style Consistency), where the former is to screen and adjust the position of foreground objects, while the latter is for the rationality of referring expressions. ComCOCO achieves a well-balanced distribution of categories and positions as depicted in~\cref{fig:short}, and a collection of image pairs in diverse scenarios is presented in~\cref{fig:onecol11}.
	
	We conduct a language baseline to quantitatively assess reliance of models on artifacts. \Cref{tab:example18}  demonstrates that the segmentation performance drops significantly with the change of input text, similar to that observed in public datasets, which verifies minimal artifacts in ComCOCO. To subjectively evaluate the realism of the synthetic dataset, we further undertake a user study employing ComCOCO shuffled with real-world images, and 30 raters are invited to assess the authenticity of given images and point out any unreal objects if present. Table~\ref{tab:example17} affirms that ComCOCO effectively reduces the visual disparities with real-world images, a testament to the rigorous manual inspection process employed during dataset construction.
	\begin{table}[h]
			\caption{Language baseline with different text input.}
		\label{tab:example18}
		\setlength{\tabcolsep}{3pt}
		\centering
		\begin{tabular}{cccccc}
			\toprule
			\multirow{2}{*}{Input} &\multicolumn{2}{c}{LAVT~\cite{Yang_2022_CVPR}}&\multicolumn{2}{c}{SAHM$_{Seg}$} &\\  \cmidrule(r){2-3} \cmidrule(r){4-5}  
			&\small{ComCOCO}&\small{RefCOCO}&\small{ComCOCO}&\small{RefCOCO}\\ 
			
			\midrule
			\small{Original Text}&64.99 & 72.73 &68.15 &72.02\\
			
			\small{Random Text}& 3.15&5.67 &3.22&2.08\\
			\small{Null Text} & 4.29& 3.03 & 5.86&3.74\\
			\bottomrule
		\end{tabular}
	
	\end{table}
	
	\begin{table}[h]
		\caption{User study of realism of ComCOCO. ($\cdot$) refers to the accuracy of pointing out unreal objects. }
		\label{tab:example17}
		\setlength{\tabcolsep}{5pt}
		\centering
		\begin{tabular}{cccccc}
			\toprule
			Dataset & Real&Slightly Unreal&Obviously Unreal \\ 
			
			\midrule
			ComCOCO&84.2\% & 15.4\% (73.8\%)  &0.4\% (95\%)\\
			RefCOCO & 98.9\%& 1.1\% & 0.0\%\\
			Places& 99.2\%&0.7\% &0.1\%\\
			\bottomrule
		\end{tabular}
		
	\end{table}
	
	\textbf{ComCOCO with spatial attributes.} The ComCOCO dataset originates from RefCOCO+ which is primarily centered around appearance-based attributes. Nevertheless, a significant drawback of the automated construction process lies in the absence of spatial attributes. This issue is further elaborated upon in the appendix. Spatial properties of objects can be divided into two categories: object-based sentences (e.g. ``\textit{the man against the door}'') and image-based sentences (e.g. ``\textit{the left truck}''). Following \cref{3.1}, for object-based sentences, all mentioned objects are expected to be reasonably positioned in the new scene with relative position invariance. However, the placement of multiple objects is prone to apparent contradiction and is laborious to discover the appropriate new scene. For image-based sentences, restricting the placement region of objects according to original descriptions is impractical since rationality is not guaranteed once the placement position is artificially constrained.
	Considering the various obstacles of automatic generation, we do not seek consistency in placement and description, and manually supplement each object with spatial attributes in the synthetic image instead. Specifically, professional annotators will describe the spatial relationship of each object with other entities and the whole image in ComCOCO, which is synchronously completed in the second manual check after Step 3 (Style Consistency).  This method ensures the accuracy of spatial information while reducing manpower costs.  
	
	To maintain independence between appearance-based expressions and viewer perspective, we intend to release two separate versions of ComCOCO, namely ComCOCO and ComCOCO+, to cater to distinct application scenarios. The primary distinction lies in the inclusion of spatial attributes in ComCOCO with 136,495 expressions, while ComCOCO+ with 107,912 expressions focuses solely on appearance-based descriptions. Both versions have comparable statistics except for the additional referring expressions of spatial attributes. This design choice facilitates effective comparison and evaluation. In this paper, we experiment on the ComCOCO version that incorporates all attributes.
	
	\section{Qualitative Analysis}\label{appendixC}
\begin{figure*}[h]
		\centering
		\includegraphics[width=1.0\linewidth]{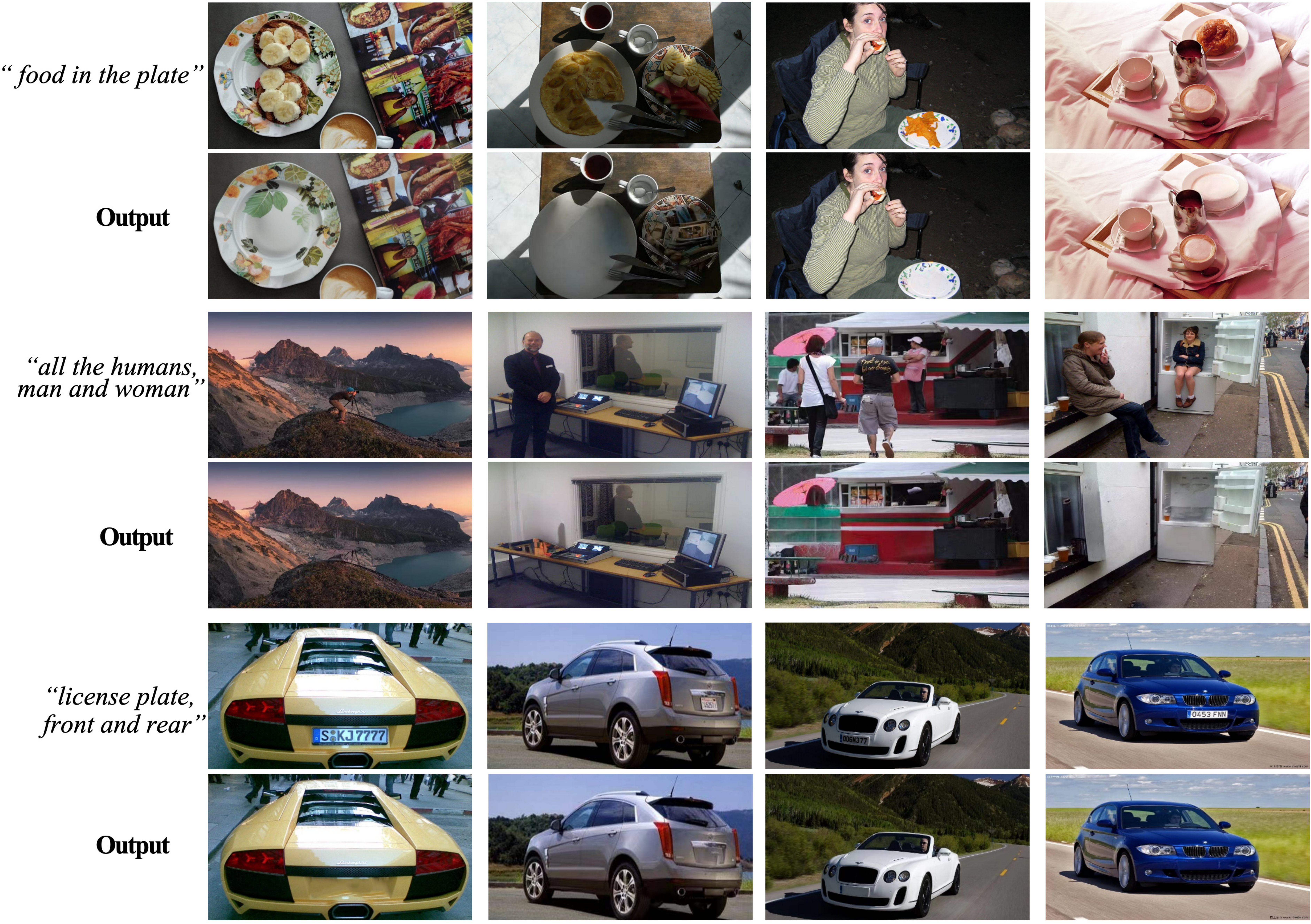}
		\caption{Removal results of our model in massive images. }
		\label{fig:onecol20}
	\end{figure*}

\begin{figure*}[h]
		\centering
		\includegraphics[width=1.0\linewidth]{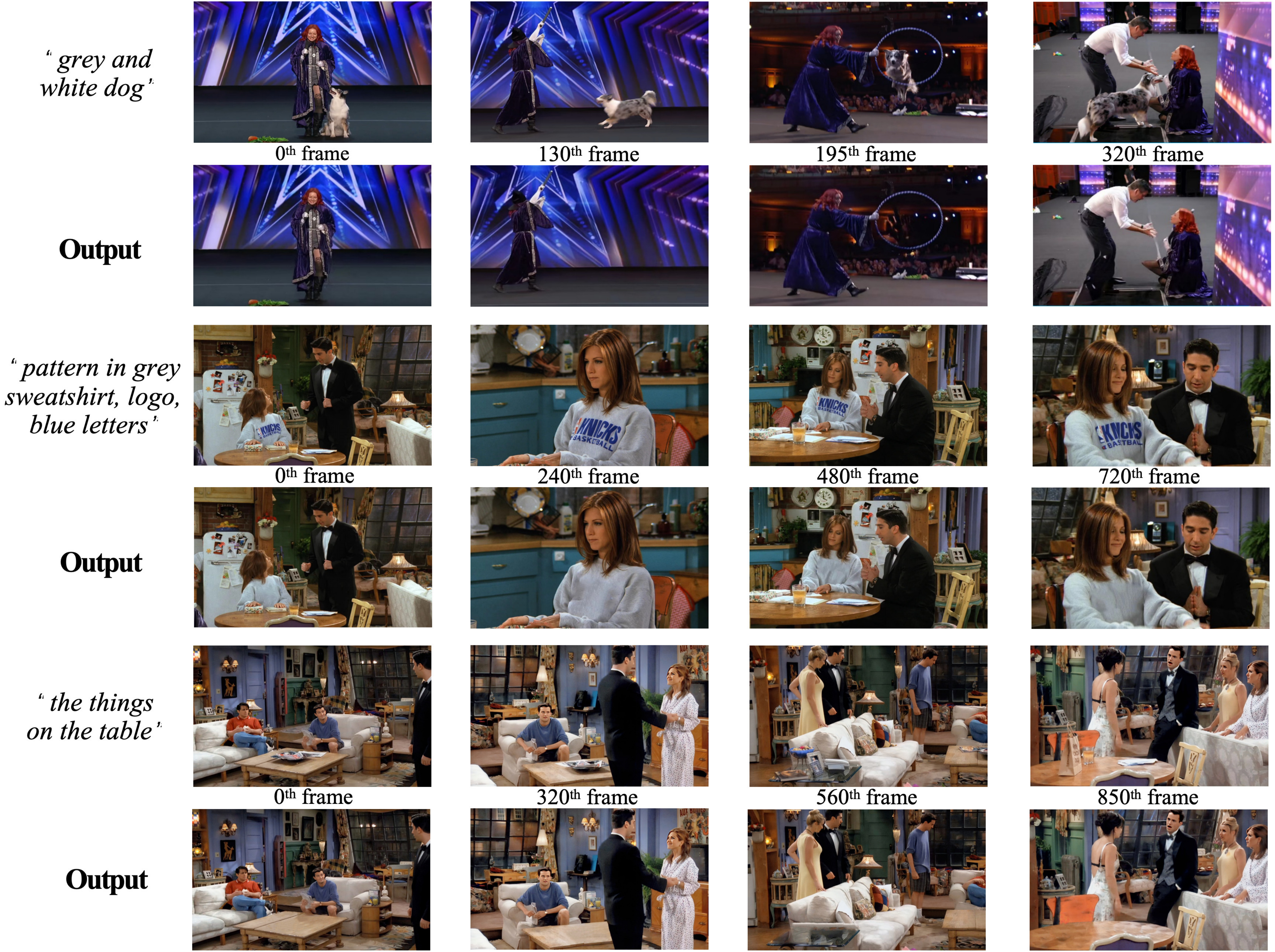}
		\caption{Removal results of our model in dynamic videos. }
		\label{fig:onecol230}
	\end{figure*}

	\textbf{Real-world Applications.} The primary application of ROR is performing similar manipulations on extensive datasets. (1) Massive images in \cref{fig:onecol20}. A typical example is image desensitization that involves removing human faces, license plates,	and other private data, particularly in autopilot technology. (2) Dynamic videos in \cref{fig:onecol230}. Specific individuals and branded products need to be removed according to different commands driven by commercial interests and public relations, especially in TV programs. In the aforementioned real-world scenarios, existing methods rely on manual pixelation of objects, which is time-consuming and introduces poor visual experience. .

\begin{figure*}[h]
		\centering
		\includegraphics[width=0.98\linewidth]{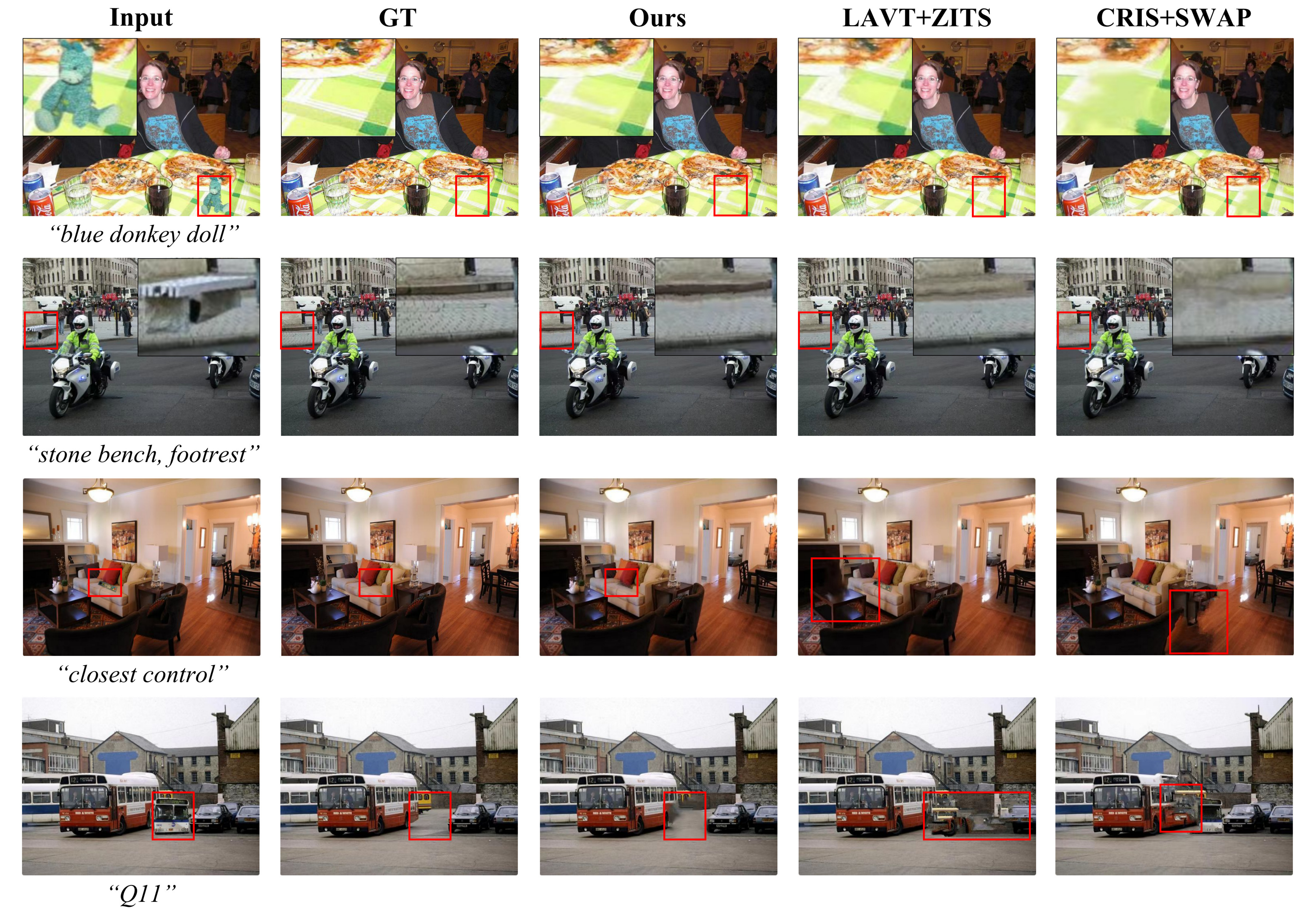}
		
		\caption{Comparison with other combined methods in ComCOCO. The first two lines compare the effectiveness of inpainting under the same location circumstance, and the last two lines exhibit the difference in locating ability.}
		\label{fig:onecol14}
	\end{figure*}

\begin{figure*}[h]
		\centering
		\includegraphics[width=0.98\linewidth]{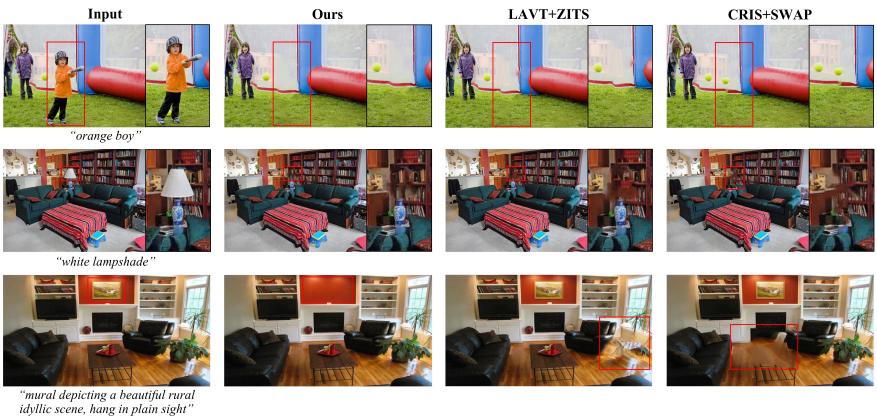}
		
		\caption{Comparison with two-stage methods in realistic images. }
		\label{fig:onecol15}
	\end{figure*}

	\textbf{Qualitative Comparison.} We visually compare our method with other combined two-stage methods in ComCOCO (\cref{fig:onecol14}) and realistic images (\cref{fig:onecol15}). Our approach can precisely analyze syntactic components and locate referred objects with oversimplified or complex descriptions. The removal results are closer to the ground truth and more convincing with visual observation.
	
	\begin{table}[h]
		\caption{Clusters for 80 object categories in RefCOCO+. X\_p refers to the cluster which includes only a part of X, and the rest of X pertains to other clusters.}
		\label{tab:example9}
		\centering
		\renewcommand\arraystretch{1.5}
		\begin{tabular}{c|c|l}
			\toprule
			\textbf{Index}& \textbf{Cluster} &\multicolumn{1}{c}{\textbf{Category}}   \\
			\midrule
			0     & person & person                                                                                                                         \\\cline{1-3}
			1     & traffic\_p             & bicycle, car, motorcycle, bus, truck                                                                                           \\ \cline{1-3}
			2     & airplane              & airplane                                                                                                                       \\\cline{1-3}
			3     & train                 & train                                                                                                                          \\\cline{1-3}
			4     & boat                  & boat                                                                                                                           \\\cline{1-3}
			5     & municipal engineering & traffic light, fire hydrant, stop sign, parking meter, bench                                                                   \\\cline{1-3}
			6     & animal                &\begin{tabular}[c]{@{}l@{}}  bird, cat, dog, horse, sheep, cow, elephant, bear\\ zebra, giraffe           \end{tabular}                                                       \\\cline{1-3}
			7     & belongings            & backpack, umbrella, handbag, tie, suitcase                                                                                      \\\cline{1-3}
			8     & sport\_p               &\begin{tabular}[c]{@{}l@{}} frisbee, sports ball, kite, baseball bat, baseball glove\\ skateboard, tennis racket    \end{tabular}                                      \\\cline{1-3}
			9     & snow   sports         & skis, snowboard                                                                                                                 \\\cline{1-3}
			10    & surfboard              & surfboard                                                                                                                      \\\cline{1-3}
			11    & tableware             & bottle, wine glass, cup, fork, knife, spoon, bowl                                                                               \\ \cline{1-3}
			12    & natural foods         & banana, apple, sandwich, orange, broccoli, carrot                                                                              \\ \cline{1-3}
			13    & processed   foods     & hot dog,   pizza, donut, cake                                                                                                  \\ \cline{1-3}
			14    & large   furniture     & chair,   couch, bed, dining table, toilet, refrigerator                                                                                          \\ \cline{1-3}
			15    & middle   furniture    & tv,   microwave, oven, toaster, sink                                                                                          \\ \cline{1-3}
			16    & miniature furniture   &\begin{tabular}[c]{@{}l@{}}potted   plant, laptop, mouse, remote, keyboard, cell phone,  \\ book, clock, vase,   scissors, teddy bear, hair drier, toothbrush\end{tabular}\\ 
			\bottomrule
		\end{tabular}
	\end{table}
	
	\captionsetup[subfigure]{singlelinecheck=off, justification=raggedleft}
\end{appendix}
\end{document}